\pdfoutput=1

\documentclass[11pt]{article}

\usepackage[final]{acl}

\usepackage{times}
\usepackage{latexsym}

\usepackage[T1]{fontenc}

\usepackage[utf8]{inputenc}

\usepackage{microtype}

\usepackage{inconsolata}

\usepackage{amsmath}
\usepackage{inconsolata}
\usepackage{amsmath}
\usepackage{graphicx}
\usepackage{multicol}
\usepackage{multirow}
\usepackage{booktabs}
\usepackage{amssymb}
\usepackage{enumitem}
\usepackage{tabularx}
\usepackage{booktabs}
\usepackage{marvosym}
\usepackage[ruled,linesnumbered]{algorithm2e}

\SetKwInput{KwIn}{Data}
\SetKwInput{KwOut}{Result}
\SetKwInput{KwReturn}{Return}
\SetKwInput{KwParameter}{Parameter}

\makeatletter
\DeclareRobustCommand\onedot{\futurelet\@let@token\@onedot}
\def\@onedot{\ifx\@let@token.\else.\null\fi\xspace}

\def\eg{\emph{e.g}\onedot} 
\def\ie{\emph{i.e}\onedot} 
 
\newcommand{\ourmethod}[1]{}
\renewcommand{\ourmethod}[1]{\textsc{DELL}}

\title{\ourmethod{}: Generating Reactions and Explanations \\for LLM-Based Misinformation Detection}

\author{
Herun Wan$^{*}$\textsuperscript{1}\ \ \ \ \ \ \
Shangbin Feng$^{*}$\textsuperscript{2}\ \ \ \ \ \ \
Zhaoxuan Tan\textsuperscript{3}\\ \bf
Heng Wang\textsuperscript{1}\ \ \ \ \ \ \ 
Yulia Tsvetkov\textsuperscript{2}\ \ \ \ \ \ \
Minnan Luo\textsuperscript{\Letter 1} \\ 
\textsuperscript{1} School of Computer Science and Technology,\\ Xi’an Jiaotong University, Xi’an, 710049, China\\
\textsuperscript{2} University of Washington \ \ \ \textsuperscript{3} University of Notre Dame \\
\href{mailto:wanherun@stu.xjtu.edu.cn}{\texttt{wanherun@stu.xjtu.edu.cn}};\ \href{mailto:shangbin@cs.washington.edu}{\texttt{shangbin@cs.washington.edu}}
}

\begin{document}
\maketitle
\def\thefootnote{*}\footnotetext{These authors contributed equally.}\def\thefootnote{\arabic{footnote}}
\def\thefootnote{\Letter}\footnotetext{Corresponding Author: \href{mailto:minnluo@xjtu.edu.cn}{\texttt{minnluo@xjtu.edu.cn}}}\def\thefootnote{\arabic{footnote}}

\begin{abstract}
Large language models are limited by challenges in factuality and hallucinations to be directly employed off-the-shelf for judging the veracity of news articles, where factual accuracy is paramount. In this work, we propose \ourmethod{} that identifies three key stages in misinformation detection where LLMs could be incorporated as part of the pipeline: 1) LLMs could \emph{generate news reactions} to represent diverse perspectives and simulate user-news interaction networks; 2) LLMs could \emph{generate explanations} for proxy tasks (\eg, sentiment, stance) to enrich the contexts of news articles and produce experts specializing in various aspects of news understanding; 3) LLMs could \emph{merge task-specific experts} and provide an overall prediction by incorporating the predictions and confidence scores of varying experts. Extensive experiments on seven datasets with three LLMs demonstrate that \ourmethod{} outperforms state-of-the-art baselines by up to 16.8\% in macro f1-score. Further analysis reveals that the generated reactions and explanations are greatly helpful in misinformation detection, while our proposed LLM-guided merging helps produce better-calibrated predictions. \footnote{Available at \url{https://github.com/whr000001/DELL}.}
\end{abstract}

\begin{figure*}
    \centering
    \includegraphics[width=\linewidth]{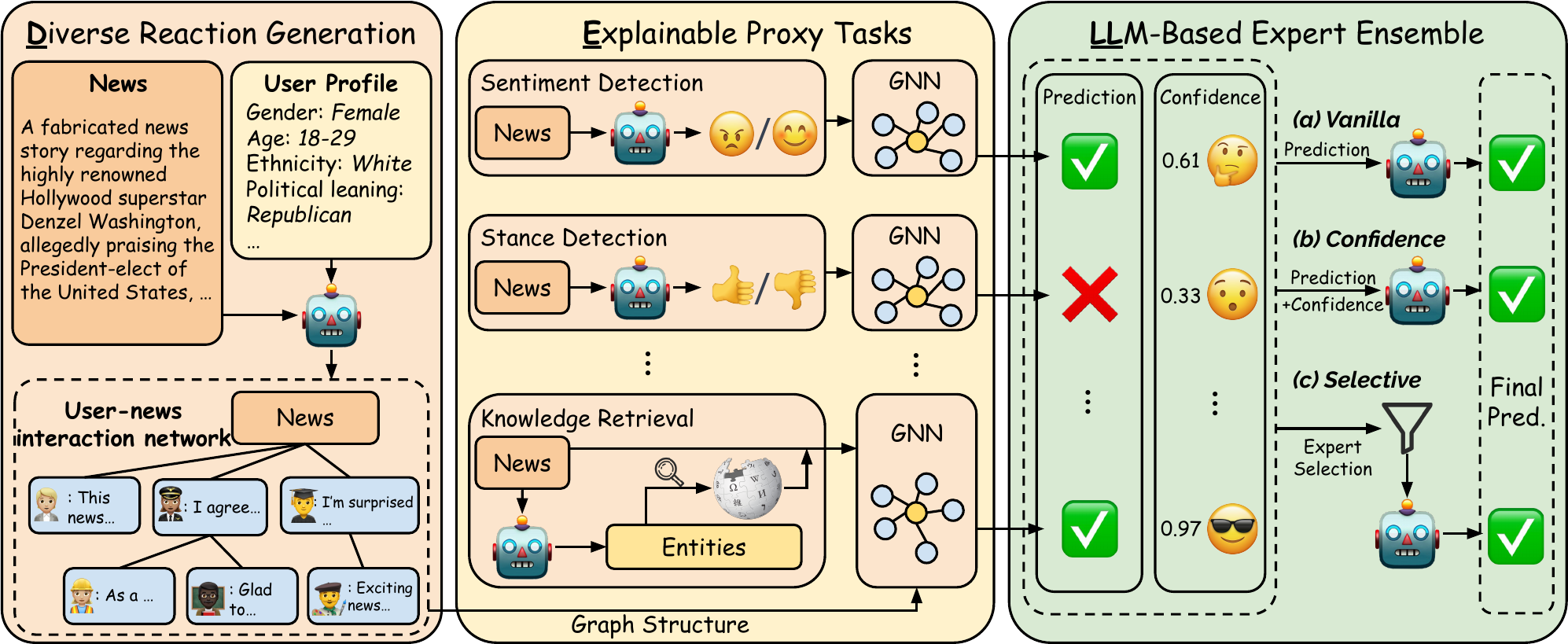}
    \caption{Overview of \ourmethod{}. We first employ LLMs to generate news reactions from diverse perspectives and form user-news interaction networks. We then design six explainable proxy tasks to refine the feature embeddings with LLM-generated explanations. We finally propose three LLM-based strategies to selectively merge the predictions of task-specific experts and enhance calibration.}
    \label{fig: overview}
    \vspace{-15pt}
\end{figure*}

\section{Introduction}
Large language models (LLMs) have demonstrated impressive capabilities to follow instructions \citep{ouyang2022training}, perform knowledge-intensive tasks \citep{rubin2022learning, shi2023replug}, and confront societal challenges \citep{jiang2023raucg, roy2023probing}. However, LLMs are also hindered by hallucinations \citep{kryscinski2020evaluating, pagnoni2021understanding, dong2022survey}, lack of factuality \citep{kandpal2023large, mallen2023not}, and challenges to adapt to new knowledge  \citep{de2021editing, hase2021language}. Despite preliminary efforts \citep{chen2023can,lucas2023fighting}, LLMs cannot yet be employed off-the-shelf for analyzing the veracity of news articles where factual accuracy is paramount \citep{leite2023detecting, hu2023bad}. Together with emerging risks of generating misinformation at scale \citep{chen2023can, wu2023fake}, these limitations call for new solutions to leverage LLMs to counter online fake news and misinformation campaigns. 

While LLMs are not reliable in detecting misinformation directly, we propose \textbf{\ourmethod{}}\footnote{\ourmethod{} stands for \textbf{\underline{D}}iverse Reaction Generation; \textbf{\underline{E}}xplainable Proxy Tasks; and \textbf{\underline{LL}}M-Based Expert Ensemble.}, employing three key stages where LLMs could be integrated to provide more context and explanations for reliable assessment of news veracity (Figure \ref{fig: overview}):

\begin{itemize}[leftmargin=*]
    \item Community reactions and comments to news articles have been shown to improve misinformation detection systems  \citep{grover2022public}. However, it is not always feasible to collect sufficient volumes of real-time user interactions \citep{he2023reinforcement}. Guided by LLMs' potential in simulating human samples and populations \citep{argyle2023out}, we employ LLMs to \textbf{generate synthetic reactions and comments} to news articles from diverse perspectives, turning the news context into a rich network of user-news interactions.
    \item Previous research shows that additional pragmatic contexts such as sentiment and stance, as well as external knowledge help aid misinformation detection \citep{zhang2021mining, hu2021compare, sengan2023fake}. To this end, we employ LLMs for \textbf{proxy tasks}, \ie, tasks where predictions and explanations could be helpful to better understand the news article. For example, LLMs evaluate the sentiment of news articles and generate both predictions and explanations: these additional contexts are then encoded as initial embeddings in the user-news interaction network for classification based on graph neural networks (GNNs). By employing six proxy tasks focusing on the news article and generated reactions, we obtain a suite of specialized predictors that specialize in various aspects of news understanding.
    \item Finally, we employ LLMs as judges to \textbf{merge the task-specific experts} and predict the news veracity. Since not all experts are equally helpful/confident for a given news article, we provide LLMs with the predictions and confidence scores of experts specializing in each proxy task: the LLM is then instructed to selectively incorporate the predictions of experts for an overall decision.
\end{itemize}

We conduct extensive experiments to evaluate \ourmethod{} and state-of-the-art baselines with three LLMs on seven datasets spanning three tasks related to news veracity, featuring both human-written and machine-generated misinformation. \ourmethod{} outperforms the strongest baseline across all datasets, achieving an improvement of up to 16.8\% in macro f1-score. Further analysis reveals that LLM-generated news reactions and explanations to proxy tasks contribute greatly to model performance, while the LLM-guided expert merging results in better-calibrated misinformation detectors for both human- and machine-generated news.

\section{Methodology}
We propose three strategies to integrate LLMs in evaluating news veracity: (\romannumeral 1)  \textbf{Diverse Reaction Generation}, leveraging LLMs to generate synthetic news reactions from diverse perspectives and forming networks of user-news interactions; (\romannumeral 2) \textbf{Explainable Proxy Tasks}, enriching news contexts and refining node embeddings in user-news interaction networks with LLM-generated task explanations; (\romannumeral 3) \textbf{LLM-Based Expert Ensemble}, adopting LLMs to selectively merge the predictions of task-specific experts and enhance calibration.

\subsection{Diverse Reaction Generation}
\label{sec: diverse_user}
Integrating the public discourse to evaluate news veracity is widely employed to better ground news articles and provide more context \citep{grover2022public, sheng2022zoom, wu2023decor, shovon2023performance}. However, real-world comments and reactions are challenging to collect, while malicious comments aiming to bolster misinformation might be removed from social media platforms and hinder reproducible research \citep{jung2020caution, grover2022public, he2023reinforcement}. Motivated by LLMs' successes in simulating human samples \citep{argyle2023out} and reflecting diverse perspectives \citep{sorensen2023value}, we propose to generate synthetic comments and reactions by LLMs, simulating how populations from diverse perspectives might respond to news articles.

\paragraph{Diverse User Attribute}
We first define the space of social media user attributes to simulate. Specifically, each synthetic user is represented as an intersection of seven categories:\footnote{We select these categories and attributes from \href{https://www.pewresearch.org/methods/2019/02/27/growing-and-improving-pew-research-centers-american-trends-panel/}{The Pew Research Center’s American Trends Panel}. Full list of potential attributes and example prompts in Appendix \ref{app: user_attribute}.} \textit{gender}, \textit{age}, \textit{ethnicity}, \textit{education}, \textit{family income}, \textit{political leaning}, and \textit{voter registration}. 
Formally, for a user attribute $\boldsymbol{P}_i$ $(1\leq i\leq n, n = 7)$, its candidate set is $\{p_i^j\}_{j=1}^{n_i}$ where $n_i$ denote the number of possibilities for a given attribute category. We sample uniformly for each user attribute to represent a social media user. We then verbalize these attributes and concatenate them as the prompt $\boldsymbol{u}$ for the synthetic user.

\paragraph{Generating User-News Networks}
Aside from news content, the non-sequential propagation structure of news comments is shown to aid in evaluating news veracity \citep{ma2018rumor, lu2020gcan, ma2023propagation}. Formally, given a news article $\boldsymbol{s}$, we aim to generate a user-news interaction network $\mathcal{G}(\mathcal{V}, \mathcal{E})$, where $\mathcal{V}$ and $\mathcal{E}$ denote the node and edge sets. We develop three strategies for LLMs to simulate the comment propagation process: (\romannumeral 1) generate a comment on the news article; (\romannumeral 2) generate a comment on a given comment; and (\romannumeral 3) select a comment to engage:

\begin{itemize}[leftmargin=*]
    \item \textbf{Comment on news}. We first generate a synthetic user description $\boldsymbol{u}$ (\S\ref{sec: diverse_user}) and append the following prompt: ``\textit{You view a piece of news with the following content. News:} $\boldsymbol{s}$''. The LLM is then instructed to generate a comment representing the user's perspective, specifically with the prompt ``\textit{Please comment on this news on social media.}''
    \item \textbf{Comment on a comment}. Similarly, we first provide LLMs with the user description $\boldsymbol{u}$ and news article $\boldsymbol{s}$. We append a comment chain $\boldsymbol{C}=[\boldsymbol{c}_1\|\boldsymbol{c}_2\|\dots\|\boldsymbol{c}_m]$, where $\boldsymbol{c}_i$ is a comment on $\boldsymbol{c}_{i-1}$. The LLM is then instructed to generate a comment to the last comment with ``\textit{Please reply to the last comment.}''
    \item \textbf{Select a comment to comment}. Social media users would selectively engage with certain comments informed by their perspectives. We employ LLMs to simulate this process by appending $\boldsymbol{u}$, $\boldsymbol{s}$, and multiple comment chains $\boldsymbol{C}$, while instructing the LLM with ``\textit{Please select a comment chain that you would most like to reply.}''
\end{itemize}

We iteratively adopt these prompts to generate a user-news interaction network for a given news article. Algorithm \ref{algo: networks} in Appendix \ref{app: user-news_networks} presents details on the user-news network generation process. 

\subsection{Explainable Proxy Tasks}
\label{sec: proxy}
Integrating LLM-generated contexts about a given document has proven effective in analyzing text-attribute graphs such as scholarly networks \citep{he2023harnessing, chen2023exploring, li2023survey}. In the domain of misinformation detection, there is often much implied context that goes beyond the news text itself, such as author stances, sentiment, external knowledge, and more. We propose to employ LLM-generated explanations for proxy tasks, \ie, tasks that help evaluate news veracity, enriching news contexts and refining the feature embeddings of user-news interaction networks with the generated explanations. Specifically, we propose four proxy tasks to enhance news articles:

\begin{itemize}[leftmargin=*]
    \item \textbf{Sentiment Analysis} News articles often feature sentiment signals that are indicative of their veracity \citep{zhang2021mining}. We employ six basic emotions \citep{ekman1999basic} (\eg, \emph{anger} and \emph{surprise}) and prompt LLMs to choose the three most likely emotions and provide explanations.
    \item \textbf{Framing Detection} Framing is a strategic device in political communication \citep{framing} and has been an integral part of evaluating news veracity \citep{kwak2020systematic,mendelsohn2021modeling}. Similarly, we follow the taxonomy of 14 media frames \citep{card2015media} (\eg, \emph{economic}) and prompt LLMs to choose the five most likely media frames and provide explanations.
    \item \textbf{Propaganda Tactics Detection} Propaganda tactics are employed to influence people's mindsets to advance a specific agenda \citep{glowacki2018news}. We follow the taxonomy of 19 propaganda tactics \citep{piskorski2023semeval} (\eg, \emph{doubt} and \emph{red herring}) and employ LLMs to identify the underlying tactics in news articles with explanations.
    \item \textbf{Knowledge Retrieval} Retrieval-augmented language models \citep{borgeaud2022improving, shi2023replug, asai2023self, chen2023dense} have demonstrated impressive potential to expand the knowledge access of LLMs. We employ LLMs to identify key entities in a news article through prompting and retrieve Wikipedia passages about these entities\footnote{We employ the \href{https://pypi.org/project/wikipedia/}{Wikipedia API} for retrieval.}. We prepend the retrieved external knowledge in the news article to facilitate better contextual understanding.
\end{itemize}

Besides news content, we also propose two proxy tasks to enhance the generated comments:

\begin{itemize}
    \item \textbf{Stance Detection} Given two text nodes $\boldsymbol{s}_1$ and $\boldsymbol{s}_2$ (news or comments) that are connected in the user-news interaction network $\mathcal{G}$, we employ LLMs to evaluate whether $\boldsymbol{s}_1$ and $\boldsymbol{s}_2$ are \emph{supportive}, \emph{neutral}, or \emph{opposed} to each other with explanations.
    \item \textbf{Response Characterization} Given two text nodes $\boldsymbol{s}_1$ and $\boldsymbol{s}_2$ (news or comments) in $\mathcal{G}$, we employ LLMs to analyze whether one is in response to another. The generated explanations would help better understand the propagation structure of news and comments.
\end{itemize}

By employing any of the six proxy tasks\footnote{We provide the prompts for proxy tasks in Appendix \ref{app: proxy}.}, we obtain an LLM-generated explanation paragraph $\boldsymbol{s}_{\textit{ext}}$ that analyzes the news article from one specialized aspect. We leverage the LLM-generated explanations to refine the feature embeddings of user-news interaction networks. Specifically, we first adopt a separate encoder-based LM $\mathrm{enc}(\cdot)$ to encode the news article $\boldsymbol{s}_{\textit{ori}}$ and the explanation $\boldsymbol{s}_{\textit{ext}}$, \ie, $\mathbf{h}_{\textit{ori(ext)}} = \mathrm{enc}(\boldsymbol{s}_{\textit{ori(ext)}})$, where we employ DeBERTa \citep{deberta} in practice. We then concatenate $\mathbf{h}_{\textit{ori}}$ and $\mathbf{h}_{\textit{ext}}$ and feed it into a linear layer to obtain initial node features $\mathbf{h}^{(0)}$.

We employ graph neural networks as the model for downstream tasks, which conduct message passing over the user-news network. Formally, suppose $\mathbf{h}_{v_i}^{(\ell)}$ is the representation of node $v_i$ at the $\ell$-th GNN layer, the feature update procedure is:
\begin{align*}
    \mathbf{h}_{v_i}^{(\ell)}=\mathop{\mathrm{Aggr}}\limits_{\forall v_j\in \mathcal{N}(v_i)}(\{\mathrm{Prop}(\mathbf{h}_{v_i}^{(\ell-1)};\mathbf{h}_{v_j}^{(\ell-1)})\}),
\end{align*}
where $\mathcal{N}(v_i)$ denotes the set of neighbors of node $v_i$, $\mathrm{Aggr}(\cdot)$ and $\mathrm{Prop}(\cdot)$ are aggregation and propagation functions, where GIN \citep{gin} is employed in practice. To obtain the graph-level representation of $\mathcal{G}$, we employ the mean pooling operator as the $\mathrm{Readout}(\cdot)$ function, \ie, 
\begin{align*}
    \mathbf{h}=\mathrm{Readout}(\{\mathbf{h}_{v_i}^{(\ell)}\}_{v_i\in\mathcal{V}}).
\end{align*}

Given a user-news network $\mathcal{G}$ and a label $y$, we compute the probability of $y$ being the correct prediction as $p(y \mid \mathcal{G}) \propto \exp(\mathrm{MLP}(\mathbf{h}))$, where $\mathrm{MLP}(\cdot)$ denotes an MLP layer. For binary classification, we optimize models using the cross-entropy loss and predict the most plausible label as $\arg \max_{y}p(y \mid \mathcal{G})$. For multi-label classification, we optimize models using the ZLPR \citep{su2022zlpr} loss and predict the label set as $\{y:p(y\mid\mathcal{G})>\lambda\}$, where $\lambda$ is a hyperparameter.

\begin{table*}[]
    \centering
    \resizebox{\textwidth}{!}{
    \begin{tabular}{l|cc|cc|cc|cc|cc|cc|cc}
    \toprule[1.5pt]
        \multirow{3}{*}{\textbf{Method}}&\multicolumn{4}{c|}{\textbf{Fake News Detection}}&\multicolumn{4}{c|}{\textbf{Framing Detection}}&\multicolumn{6}{c}{\textbf{Propaganda Tactic Detection}}\\
        &\multicolumn{2}{c|}{\textbf{Pheme}}&\multicolumn{2}{c|}{\textbf{LLM-mis}}&\multicolumn{2}{c|}{\textbf{MFC}}&\multicolumn{2}{c|}{\textbf{SemEval-23F}}&\multicolumn{2}{c|}{\textbf{Generated}}&\multicolumn{2}{c|}{\textbf{SemEval-20}}&\multicolumn{2}{c}{\textbf{SemEval-23P}}\\
        &MaF&MiF&MaF&MiF&MaF&MiF&MaF&MiF&MaF&MiF&MaF&MiF&MaF&MiF\\
    \midrule[1pt]
        \textsc{zero-shot}&.459&.460&.597&.600&.332&.346&.381&.443&.223&.233&.304&.424&.228&.379\\
        \textsc{few-shot}&.490&.500&.565&.570&.350&.395&.457&.512&.344&.358&.359&.468&.266&.424\\
        \textsc{retrieval}&.464&.470&.624&.630&.278&.334&.397&.480&.262&.267&.292&.415&.187&.309\\
        \textsc{F3} \textit{Z-CoT}&.499&.500&.566&.570&.285&.314&.370&.470&.223&.203&.302&.418&.248&.423\\
        \textsc{F3} \textit{DeF-Gen}&.410&.410&.477&.480&.319&.354&.381&.468&.284&.290&.331&.508&.259&.396\\
        \textsc{TAPE w/o graph}&.767&.770&.858&.860&.341&.482&.393&.631&.298&.326&.332&.565&.237&.583\\
        \textsc{DeBERTa}&.779&.780&.887&.890&.388&.543&.506&.672&.512&.516&.516&.609&.343&.558\\
    \midrule[1pt]
        \textsc{k-hops}&.374&.490&.421&.470&.332&.407&.362&.466&.206&.193&.350&.448&.280&.393\\
        \textsc{k-attention}&.325&.450&.407&.450&.348&.418&.413&.496&.214&.211&.310&.409&.198&.318\\
        \textsc{TAPE w/ graph}&.787&.790&.888&.890&.381&.515&.399&.623&.279&.306&.332&.598&.250&.581\\
    \midrule[1pt]
        \textsc{GCN}&.790&.790&.854&.860&.447&.566&.499&.658&.504&.496&.517&.628&.358&.547\\
        \textsc{RvNN}&.790&.790&.888&.890&.428&.551&.494&.644&.494&.496&.462&.559&.363&.568\\
        \textsc{dEFEND}&.727&.730&.823&.840&.434&.607&.435&.557&.063&.099&.280&.576&.255&.601\\
        \textsc{Hyphen}&.777&.780&.836&.840&.481&\textbf{.634}&.528&\underline{.714}&.292&.327&.347&.508&.301&.488\\
        \textsc{GET}&.788&.790&.847&.850&.445&.566&.525&.649&.250&.227&.423&.561&.361&.617\\
        \textsc{WSDMS}&.799&.800&.860&.870&.434&.597&.526&.688&.376&.419&.509&\underline{.630}&.333&.619\\
    \midrule[1pt]
        \ourmethod{} \textit{Single}&\underline{.810}&\underline{.810}&\textbf{.928}&\textbf{.930}&.458&.598&.536&.684&.543&.556&.520&.613&\underline{.376}&.631\\
        \ourmethod{} \textit{Vanilla}&\underline{.810}&\underline{.810}&\underline{.926}&\textbf{.930}&.432&.591&.528&.689&.578&\underline{.566}&.508&.611&.365&.634\\
        \ourmethod{} \textit{Confidence}&\underline{.810}&\textbf{.820}&.917&\underline{.920}&\textbf{.509}&\underline{.603}&\textbf{.572}&\textbf{.718}&\underline{.579}&.558&\underline{.523}&.624&\textbf{.386}&\underline{.643}\\
        \ourmethod{} \textit{Selective}&\textbf{.820}&\textbf{.820}&.897&.900&\underline{.488}&.581&\underline{.554}&.683&\textbf{.598}&\textbf{.577}&\textbf{.525}&\textbf{.636}&.362&\textbf{.652}\\
    \bottomrule[1.5pt]
    \end{tabular}
    }
    \vspace{-5pt}
    \caption{Performance of \ourmethod{} and baselines on seven datasets from three misinformation-related tasks. \textit{Single} indicates the best-performing single expert. ``MaF'' and ``MiF'' indicates macro- and micro-averaged f1-score. \textbf{Bold} indicates the best performance and \underline{underline} indicates the second best. \ourmethod{} outperforms state-of-the-art baselines by up to 16.8\% in macro f1-score, indicating the success of our LLM integration strategies.}
    \label{tab: main}
    \vspace{-15pt}
\end{table*}

\subsection{LLM-Based Expert Ensemble}
By adopting different proxy tasks and LLM-generated explanations, we obtain a set of experts, where each specializes in one proxy task and various aspects of news articles. To obtain an overall prediction, we propose an LLM-based expert ensemble to selectively leverage experts, their predictions, and confidence scores. We first use one sentence $\boldsymbol{d}_i$ to describe each expert, \eg, ``This expert focuses on the emotion of news.'' We then propose three modes for LLMs to merge experts\footnote{We provide prompts in Appendix \ref{app: ensemble}.}:

\paragraph{Vanilla} LLMs are first provided with news content and an instruction, \ie, ``\textit{Some experts give predictions about the news.}'' We then append the description and prediction of each expert: for an expert $e_i$ with prediction $\boldsymbol{\ell}_i$ and its description $\boldsymbol{d}_i$, the expert prompt is ``\textit{Expert} $i$\textit{:} $\boldsymbol{d}_i$\textit{. The expert predicts the label of this news is }$\boldsymbol{\ell_i}$\textit{.}'' Finally, the LLM is instructed to reason and generate a final prediction based on the experts' feedback.
\paragraph{Confidence} In \textit{Vanilla}, we assume that all experts should be equally important. However, experts could have varying levels of confidence and we take this into account by additionally providing the confidence scores. The confidence scores are obtained from the classification layer of the GNN-based model (\S\ref{sec: proxy}). We aim to improve the calibration of LLM-based expert ensemble by incorporating confidence scores of individual experts.
\paragraph{Selective} In \textit{Vanilla} and \textit{Confidence}, we assume that every news article would benefit from the input of all experts. However, this could introduce noise in the LLM reasoning process  \citep{Feng2023KnowledgeCF, zhao2023thrust}. To this end, we propose the \textit{Selective} approach, putting LLMs in charge to selectively activate experts. Specifically, we provide news content and expert descriptions, then prompt LLMs with ``\textit{To understand this news, which expert knowledge do you need?}'' We ensemble the selected experts with the \emph{Confidence} strategy to obtain the final predictions.

\section{Experiment Settings}
\paragraph{Models and Settings} We leverage \textit{Mistral-7B} \citep{jiang2023mistral}, \textit{LLaMA2-70B} \citep{touvron2023llama}, and \textit{ChatGPT} as the base LLMs. We mainly employ \textit{Mistral-7B} to generate comments and conduct proxy tasks, and \textit{ChatGPT} to ensemble experts. We set the temperature $\tau=0.6$ for \textit{Mistral-7B} and $\tau=0.1$ for \textit{ChatGPT}. We present more results from other LLMs in Appendix \ref{app: exp}.

\paragraph{Baselines} We compare \ourmethod{} with three types of state-of-the-art baselines: 1) \textbf{LLM-only}: \textsc{zero-shot}, \textsc{few-shot}, \textsc{retrieval}-augmented generation, \textsc{F3} \textit{Z-CoT} \citep{lucas2023fighting}, \textsc{F3} \textit{DeF-Gen} \citep{lucas2023fighting}, \textsc{TAPE w/o graph} \citep{he2023harnessing}, and \textsc{DeBERTa} \citep{deberta}; 2) \textbf{LLM+Graph}: \textsc{k-hops} \citep{huang2023can} and \textsc{k-attention} \citep{huang2023can}, and \textsc{TAPE w/ graph} \citep{he2023harnessing}; 3) \textbf{Graph-based}: \textsc{GCN} \citep{DBLP:conf/iclr/KipfW17}, \textsc{RvNN} \citep{ma2018rumor}, \textsc{dEFEND} \citep{shu2019defend}, \textsc{Hypehn} \citep{grover2022public}, \textsc{GET} \citep{xu2022evidence}, and \textsc{WSDMS} \citep{DBLP:conf/emnlp/YangG0L023}. We provide more details about baselines in Appendix \ref{app: baseline}.

\paragraph{Tasks and Datasets} We evaluate \ourmethod{} and baselines on three tasks related to chacterizing misinformation, \ie, 1) fake news detection: \textbf{Pheme} \citep{buntain2017automatically} and \textbf{LLM-mis} \citep{chen2023can}, which feature a binary classification setting; 2) framing detection: \textbf{MFC} \citep{card.2015} and \textbf{SemEval-23F} \citep{piskorski2023semeval}, which feature a multi-label classification setting; 3) propaganda tactic detection: \textbf{Generated} generated by \textit{ChatGPT}, \textbf{SemEval-20} \citep{DBLP:data/10/MartinoBWPN20}, and \textbf{SemEval-23P} \citep{piskorski2023semeval}, which feature a multi-label classification setting. The datasets are all in English and we provide more dataset details in Appendix \ref{app: datasets}. To evaluate the ability to evaluate machine-generated news, \textbf{LLM-mis} is an extended version of \textbf{FakeNewsNet} \citep{shu2020fakenewsnet} and \textbf{Generated} is generated by LLMs.

\begin{table*}[]
    \centering
    \resizebox{\textwidth}{!}{
    \begin{tabular}{cccc|cccccccc}
    \toprule[1.5pt]
        \multirow{2}{*}{\textbf{Metric}}&\multicolumn{3}{c|}{\textbf{Real Networks}}&\multicolumn{8}{c}{\textbf{Simulated Networks}}\\
         &Pheme&Twitter15&Twitter16&More&Pheme&LLM-mis&MFC&SemEval-23F&Generated&SemEval-20P&SemEval-23P\\
         \midrule[0.75pt]
         betweenness&0.255&0.191&0.234&0.208&0.293&0.291&0.291&0.287&0.291&0.286&0.288\\
         shortest path&2.682&1.904&1.833&2.076&2.925&2.913&2.913&2.869&2.908&2.863&2.879\\
         degree&0.764&0.945&0.962&0.821&0.400&0.399&0.399&0.408&0.402&0.416&0.410\\
         diameter&5.477&2.848&2.605&3.281&6.006&5.942&5.951&5.793&5.929&5.792&5.840\\
         \bottomrule[1.5pt]
    \end{tabular}}
    \caption{The graph indicators of the real and simulated networks. ``More'' denotes that networks are generated when $\alpha=0.8$ and $\beta=0.05$. Our generated networks are statistically similar to those in dataset Pheme as of network structure, indicating our generation strategy could stimulate the network structures similar to the real situation.}
    \label{tab: simulate}
    \vspace{-10pt}
\end{table*}

\section{Results}
We present the performance of \ourmethod{} and state-of-the-art baselines in Table \ref{tab: main}. We present more ablation study results in Table \ref{tab: ablation} in Appendix \ref{app: exp}.

\paragraph{\ourmethod{} achieves state-of-the-art performance.}
\ourmethod{} outperforms the strongest baseline on all seven benchmarks by $1.46\%$ to $16.80\%$ on macro f1-score, indicating the success of integrating LLMs in multiple stages of news veracity evaluation. We find that LLM-only in-context learning approaches struggle in performance, indicating that LLMs are limited by factuality challenges and hallucinations to evaluate the veracity of news articles.

\begin{figure}
    \centering
    \includegraphics[width=\linewidth]{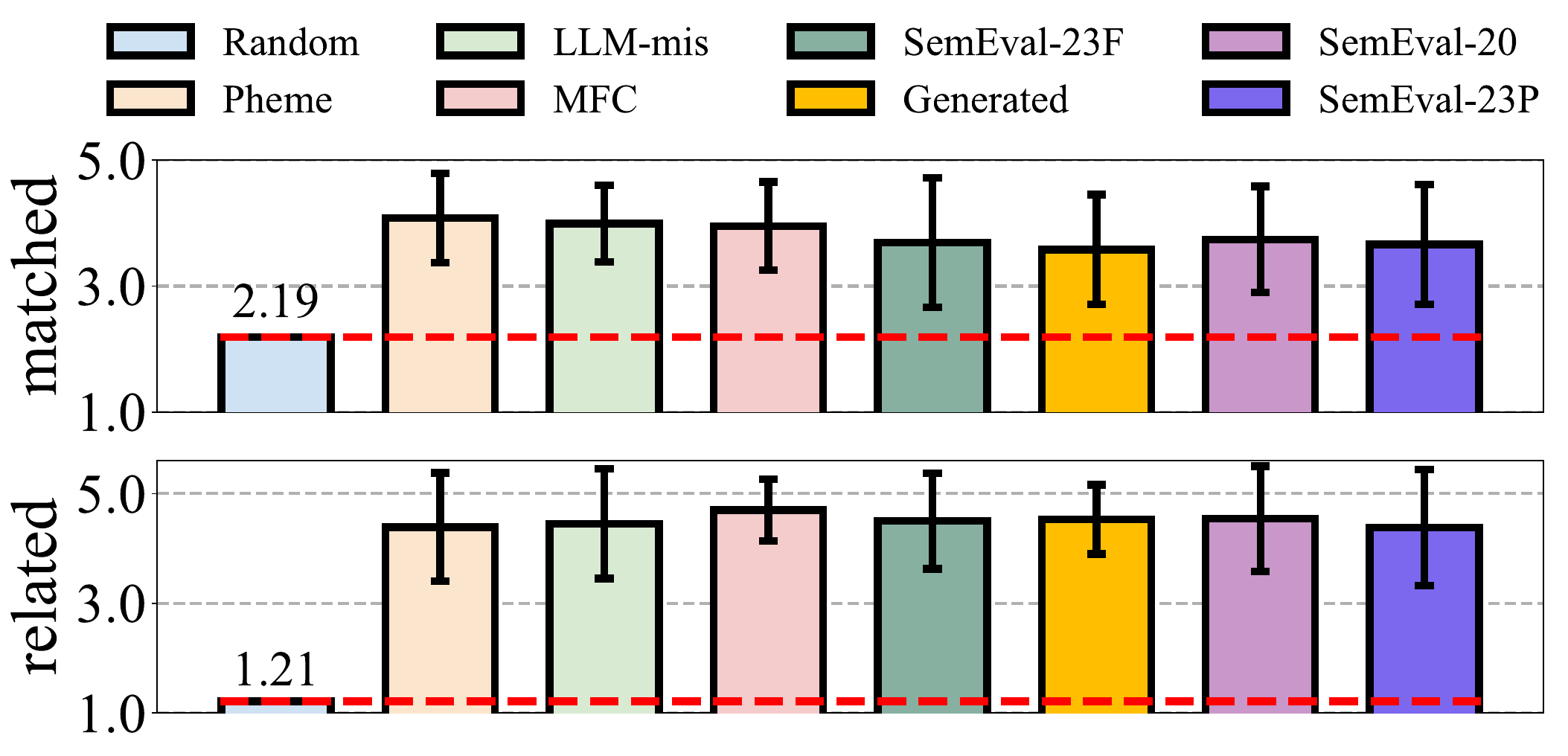}
    \caption{GPT-4 evaluation of whether the LLM-generated comments are related to the news article and match the user attributes, the higher the better from 1 to 5. We present the average value and standard deviation. Compared with randomly paired news (``Random'' in the figure), user attributes, and comments, the generated comments generally conform to the user attributes and are relevant to the news articles.}
    \label{fig: comment_quantitative}
    \vspace{-15pt}
\end{figure}

\paragraph{Generated news reactions help ground news articles.}
Compared to news-only approaches, models enhanced with generated comments (both ours and graph-based baselines) achieve better performance. The average performance on \textbf{MFC} of the comment-enhanced models is $15.2\%$ higher on MaF. It indicates that LLM-generated diverse comments are beneficial in characterizing misinformation.

\paragraph{Proxy tasks improve news understanding ability.}
\textsc{\ourmethod{}} \emph{single} denotes the performance of the best single expert focusing on one proxy task. We find that a single expert could already achieve a substantial improvement in most cases: for example, on benchmark \textbf{Generated}, it achieves a $6.16\%$ improvement on the macro f1-score than the strongest baseline. This indicates that our explainable proxy tasks are effective strategies for incorporating LLMs for evaluating news veracity.

\paragraph{LLMs could ensemble expert predictions.}
Compared to a single expert, the proposed LLM ensemble strategies achieve improvements on six out of seven datasets. In addition to simple aggregation (\textit{Vanilla}), \textit{Confidence} and \textit{Selective} improve the ensemble by accessing the confidence scores and selectively incorporating certain experts, indicating that LLMs have preliminary capabilities of understanding verbalized confidence scores \citep{tian2023just, feng2024don}. We further investigate if LLM-based ensembling could lead to better-calibrated misinformation detectors in Section \ref{sec: analysis}.

\section{Analysis}
\label{sec: analysis}
\paragraph{Quality of Generated Comments}

\begin{figure}
    \centering
    \includegraphics[width=\linewidth]{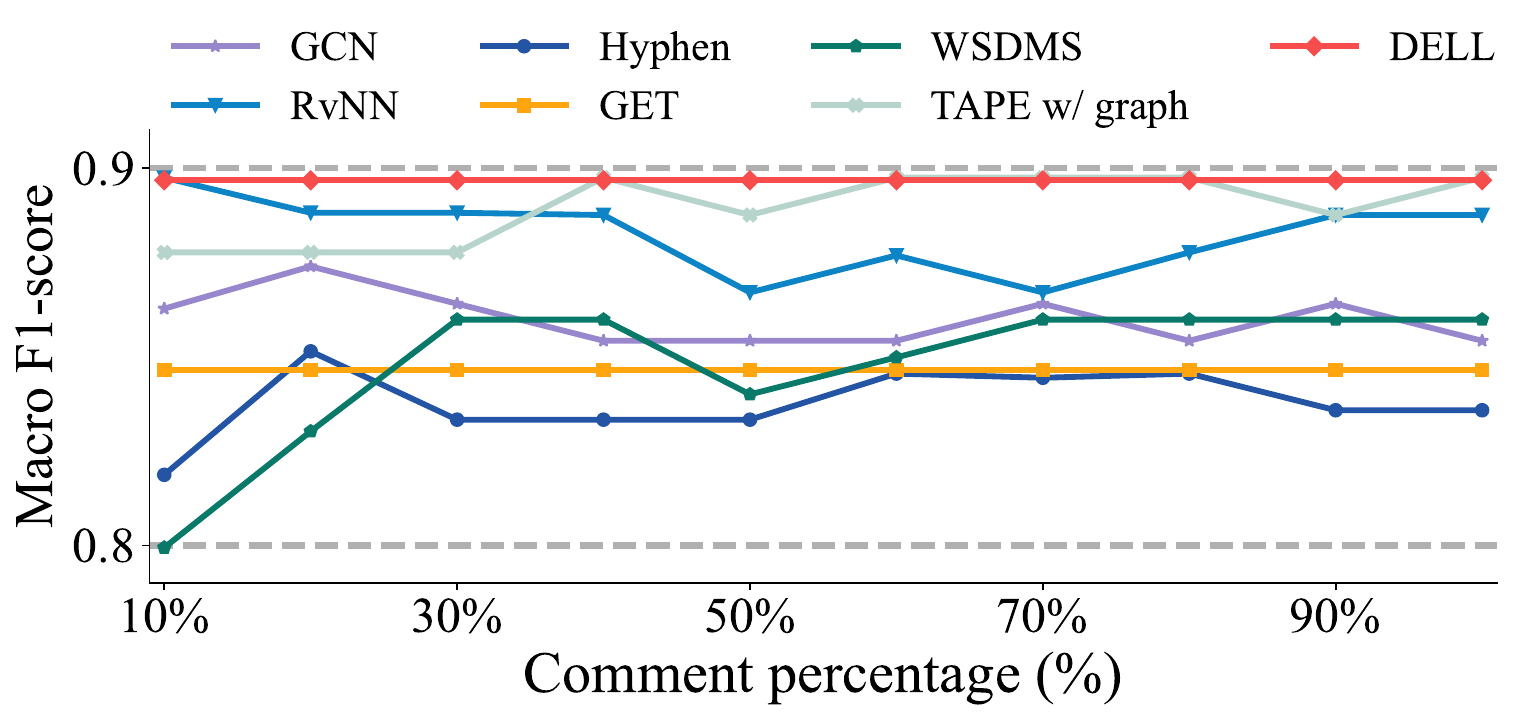}
    \caption{Performance of \ourmethod{} and baselines on \textbf{LLM-mis} when the comments are gradually removed. \ourmethod{} shows great robustness to the availability of comments.}
    \label{fig: comment_drop}
    \vspace{-15pt}
\end{figure}

We verify the quality of LLM-generated comments on whether it matches the user attributes and whether it is related to the news article. We conduct a human evaluation with four annotators to manually evaluate 50 generated comments from two datasets on a five-point Likert scale, where the higher scores mean better quality. The average score is 4.52, the standard deviation is 0.69 and the annotator agreement in Fleiss' Kappa is 0.216, which indicates that annotators generally agree that the LLM-generated comments are related and on-brand for user attributes.

We additionally employ GPT-4 evaluation \citep{chiang2023can, kim2023prometheus} for quantitative evaluation, where we randomly sample 700 generated comments and prompt GPT-4 with ``\textit{Does the user's comment on the news match the profile?}'' and ``\textit{Does the comment relate to the news?}'' to solicit a response on a five-point Likert scale. Figure \ref{fig: comment_quantitative} demonstrates that the automatic evaluation also finds that the generated comments are consistent with the user attributes and relevant to the news.

\begin{figure}
    \centering
    \includegraphics[width=\linewidth]{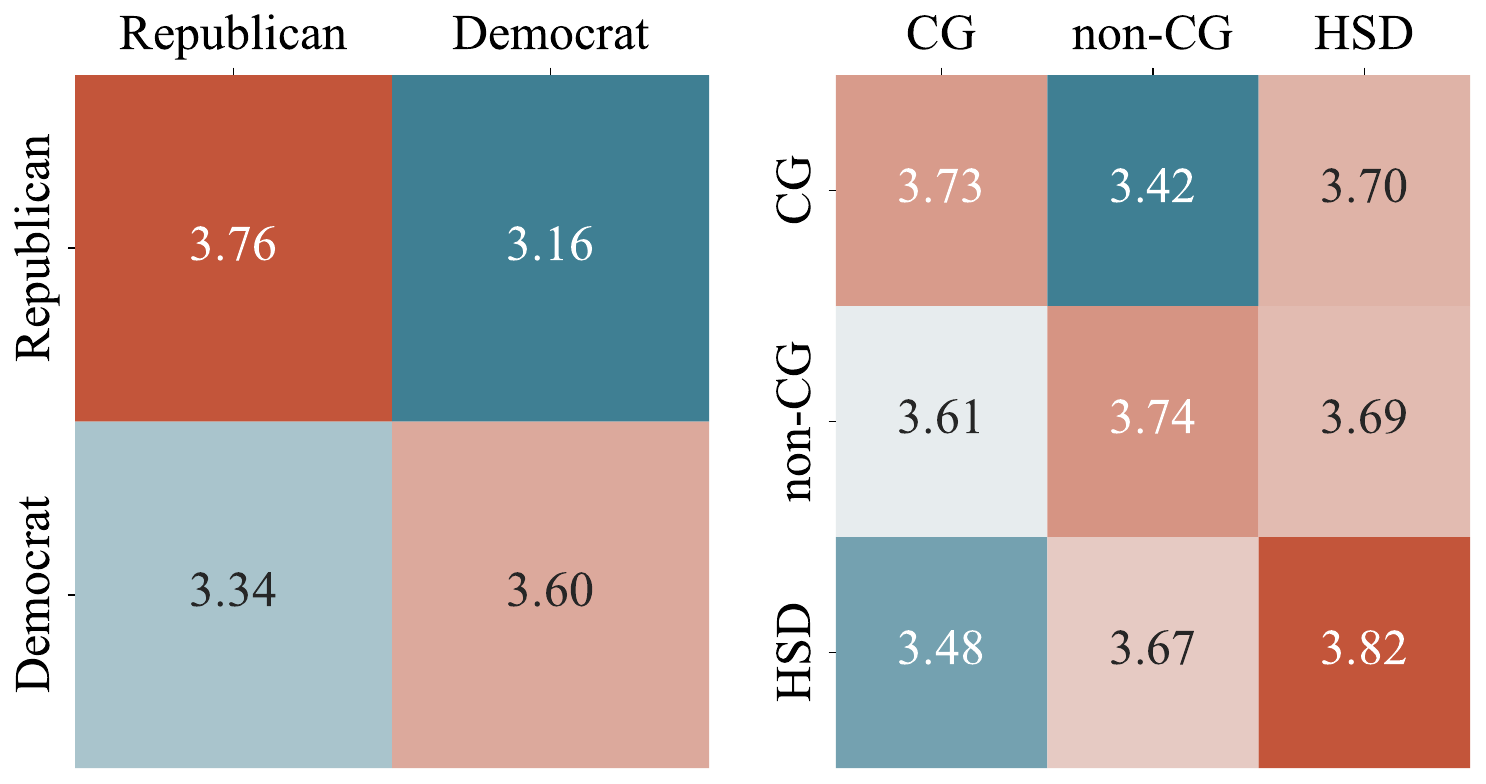}
    \caption{GPT-4 evaluation of the matching degree between different user groups. ``CG'' denotes ``college grad'', ``non-CG'' denotes ``haven't graduated from college'', and ``HSD'' denotes ``have a high school diploma or less''. The diagonal numbers are the highest both row-wise and column-wise, indicating that the generated comments are consistent with the user attributes.}
    \label{fig: heatmap}
    \vspace{-15pt}
\end{figure}

We conduct an additional evaluation to ``put a more challenging control group of comments generated by the same framework but with a different demographic'': we sample comments from users with other attributes, and then employ GPT-4 evaluation to check whether the generated comments match each attribute. For example, we sample 100 synthetic comments (50 with the attribute Democrat and 50 with the attribute Republican), and then we employ GPT4 to evaluate to what extent, on a scale of 1-5, do these comments match Democrats and Republicans. Then we could obtain 200 scores and draw a heat map. We similarly experiment with the education attribute, spanning ``college grad'', ``haven’t graduated from college'', and ``have a high school diploma or less''. In Figure \ref{fig: heatmap}, we find that the diagonal numbers, where the user attribute matches what GPT-4 evaluates, are the highest both row-wise and column-wise, indicating that the generated comments are consistent with the user attributes.

\paragraph{Network Generation Ability}
To establish that the generated interaction networks resemble real-world networks, we compare our generated networks with the real networks in datasets \textbf{Pheme} \citep{buntain2017automatically}, \textbf{Twitter-15} \citep{ma2018rumor}, and \textbf{Twitter-16} \citep{ma2018rumor}. Specifically, we calculate the average edge betweenness of each edge, the average shortest path length, the ratio of maximum degree to number of nodes, and the diameter of each graph. Then we average the value over the whole dataset to compare in Table \ref{tab: simulate}. The results show that our generated networks are statistically similar to those in dataset \textbf{Pheme} as for network structure, indicating our generation strategy could stimulate the network structures similar to the real situation. In addition, hyperparameters in Algorithm \ref{algo: networks} enable the control of generating user reaction networks. For example, by setting $\alpha=0.8$ and $\beta=0.05$, generated networks resemble those in datasets \textbf{Twitter-15} and \textbf{Twitter-16}. As a result, \ourmethod{} could reliably simulate real-world user interaction networks and structures through those control measures.

\paragraph{Model Robustness to Comments}
Since comments are usually hard to collect and generating comments using LLM could be computationally expensive, detectors should be robust to the amount of comments. We evaluate approaches on the test sets where LLM-generated comments are gradually removed. As demonstrated in Figure \ref{fig: comment_drop}, \ourmethod{} drops the least in performance with reduced comments and on dataset \textbf{LLM-mis} our performance is almost unchanged. This indicates that \ourmethod{} benefits greatly from as few as 10\% of news comments.

\begin{figure}[t]
    \centering
    \includegraphics[width=\linewidth]{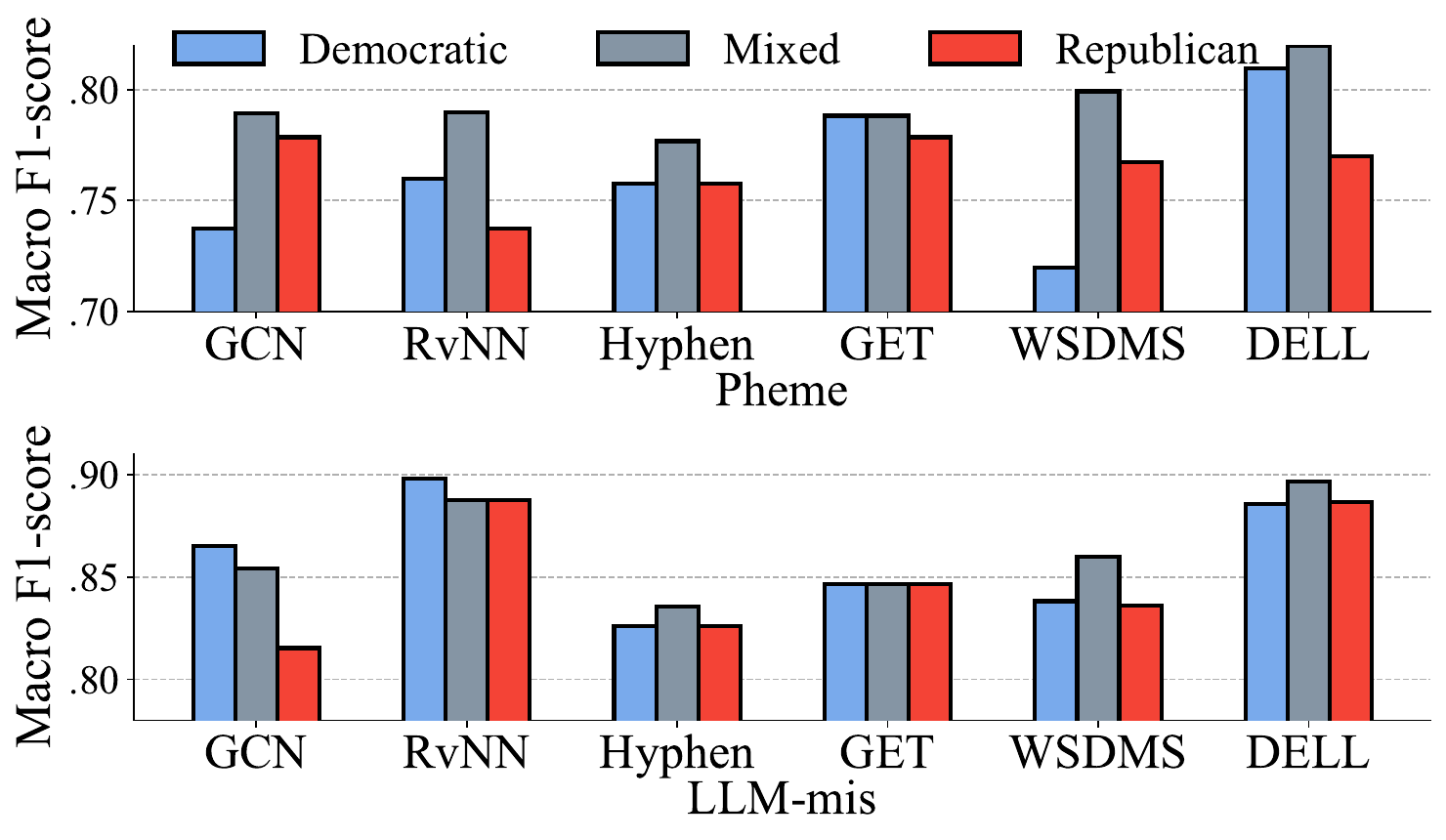}
    \caption{Performance of \ourmethod{} and baselines when the comments are generated from only one partisan perspective. Models trained on comments from one perspective generally perform worse than diverse comments.}
    \label{fig: party}
    \vspace{-10pt}
\end{figure}

\begin{table*}[]
    \centering
    \resizebox{\textwidth}{!}{
    \begin{tabular}{ll|cc|cc|ccc}
    \toprule[1.5pt]
    \multirow{2}{*}{\textbf{Strategy}}&\multirow{2}{*}{\textbf{Variants}}&\multicolumn{2}{c|}{\textbf{Fake News Detection}}&\multicolumn{2}{c|}{\textbf{Framing Detection}}&\multicolumn{3}{c}{\textbf{Propaganda Tactic Detection}}\\
    &&\textbf{Pheme}&\textbf{LLM-mis}&\textbf{MFC}&\textbf{semeval-23F}&\textbf{Generated}&\textbf{semeval-20}&\textbf{semeval-23P}\\
    \midrule[1pt]
    \multirow{3}{*}{\textit{Vanilla}}&\textbf{Original}&.810&.926&.432&.528&.578&.508&.365\\
    &\textbf{Only Content}&.799 (-1.3\%)&.885 (-4.4\%)&.446 (+3.2\%)&.537 (+1.7\%)&.570 (-1.3\%)&.520 (+2.4\%)&.397 (+8.8\%)\\
    &\textbf{Only Comments}&.780 (-3.7\%)&.927 (+0.1\%)&.449 (+4.0\%)&.533 (+1.0\%)&.436 (-24.6\%)&.526 (+3.6\%)&.345 (-5.5\%)\\
    \midrule[1pt]
    \multirow{3}{*}{\textit{Confidence}}&\textbf{Original}&.820&.917&.509&.572&.579&.523&.386\\
    &\textbf{Only Content}&.820 (+0.0\%)&.907 (-1.1\%)&.458 (-9.9\%)&.578 (+1.1\%)&.556 (-3.9\%)&.515 (-1.4\%)&.404 (+4.6\%)\\
    &\textbf{Only Comments}&.769 (-6.1\%)&.907 (-1.0\%)&.428 (-15.8\%)&.534 (-6.7\%)&.548 (-5.4\%)&.470 (-10.1\%)&.386 (-0.1\%)\\
    \midrule[1pt]
    \multirow{3}{*}{\textit{Select}}&\textbf{Original}&.820&.897&.488&.554&.598&.525&.362\\
    &\textbf{Only Content}&.800 (-2.4\%)&.907 (+1.1\%)&.477 (-2.2\%)&.540 (-2.5\%)&.579 (-3.2\%)&.526 (+0.1\%)&.360 (-0.4\%)\\
    &\textbf{Only Comments}&.770 (-6.1\%)&.917 (+2.2\%)&.426 (-12.7\%)&.547 (-1.4\%)&.529 (-11.5\%)&.507 (-3.4\%)&.394 (+8.9\%)\\
    \bottomrule[1.5pt]
    \end{tabular}
    }
    \vspace{-5pt}
    \caption{Ablation study of \textbf{expert ensemble}, where only experts of proxy tasks focusing on either news content or comments are retained. We present the macro f1-score for each variant and performance changes compared to the original setup. Diverse experts generally outperform a single type of expert, while experts who focus on news content are generally better than those who focus on comments.}
    \label{tab: ablation}
    \vspace{-10pt}
\end{table*}

\paragraph{Comment Diversity}
We propose to generate diverse comments by employing LLMs to simulate diverse user attributes. To validate this design choice, we re-generate news comments solely with synthetic Republican or Democratic users and evaluate model performance on the fake news detection benchmarks. Figure \ref{fig: party} demonstrates that only considering reactions from a single partisan viewpoint is generally worse, supporting our proposal of integrating diverse comments in fake news detection.

\paragraph{Expert Ablation}
Experts are specialized with two types of proxy tasks, focusing on either news content or comments. We conduct ablation studies to examine the impact of different types of proxy tasks. Table \ref{tab: ablation} demonstrates that: 1) integrating both types of experts leads to better performance, where the performance of a single category drops by up to 15.8\%; and 2) experts focusing on proxy tasks of news content generally outperform experts who focus solely on comments, while the two types of proxy tasks are complementary.

\begin{figure}[t]
    \centering
    \includegraphics[width=\linewidth]{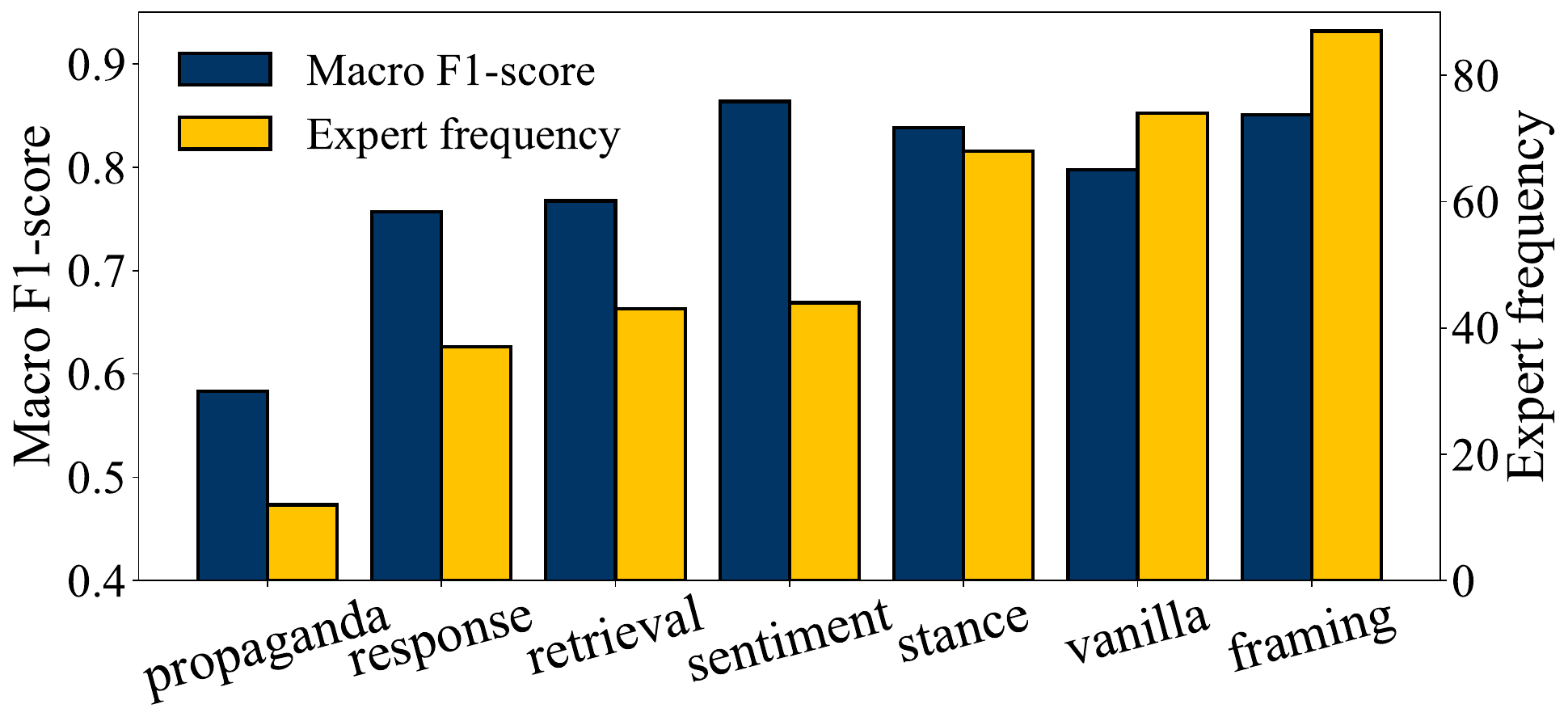}
    \vspace{-15pt}
    \caption{The frequency of expert selections and performance on \textbf{Pheme} when a particular expert is selected in the \textit{Selective} approach. Experts who have been selected more times tend to perform better.}
    \label{fig: exepert_ana}
    \vspace{-10pt}
\end{figure}

\paragraph{Expert Selection}
In the \textit{Selective} LLM-based ensemble strategy, LLMs determine which experts are activated and incorporated in the overall decision. To evaluate each expert's contribution, we examine the frequency of expert selection and the performance when a given expert is selected. Figure \ref{fig: exepert_ana} illustrates that experts who have been selected more times tend to perform better, indicating that LLMs have preliminary capabilities to select helpful experts based on the news content.

\begin{figure*}
    \centering
    \includegraphics[width=\linewidth]{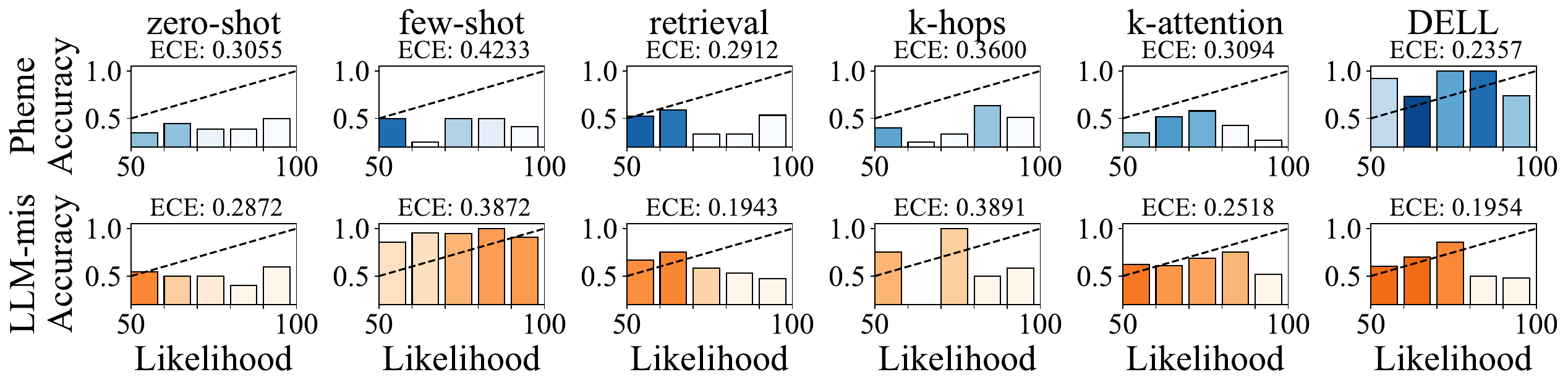}
    \vspace{-15pt}
    \caption{Calibration of \ourmethod{} and baselines in the fake news detection benchmarks. ECE denotes estimated calibration error, the lower the better. The dashed line indicates perfect calibration, while the bar's color is darker when it is closer to perfect calibration. \ourmethod{} achieves low ECE scores and thus is better-calibrated.}
    \label{fig: calibration}
    \vspace{-15pt}
\end{figure*}

\paragraph{Model Calibration}
Robust fake news detectors should provide not only a binary prediction but also a well-calibrated confidence score to facilitate content moderation. We evaluate how well \ourmethod{} and baselines are calibrated with the fake news detection datasets in Figure \ref{fig: calibration}. We use the probability of the prediction token (``fake'' or ``real'') from the LLM as the confidence score, bin it into five buckets ($0.5$ to $1.0$), and calculate the estimated calibration error (ECE) \citep{guo2017calibration}. It is demonstrated that \ourmethod{} are better-calibrated with an ECE of 0.2357 while achieving an improvement of up to 19.1\% compared to baselines. We hypothesize that by integrating expert confidence scores in the LLM-guided ensemble, the overall decision is better-calibrated and thus more trustworthy.

\begin{figure}[t]
    \centering
    \includegraphics[width=\linewidth]{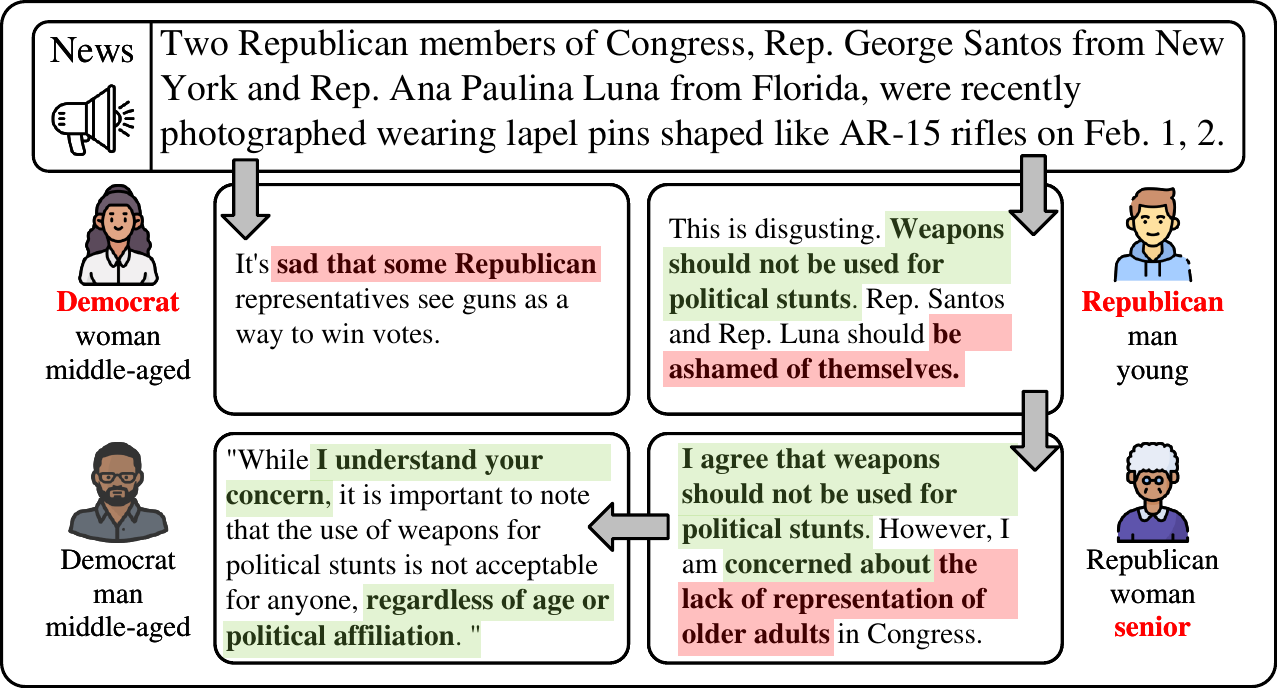}
    \caption{An example of generated comments related to a news article. The red areas indicate the comments match the user profiles, while the green areas indicate strong continuity in the comments of the comment chain. This case highlights the quality and contribution of the LLM-generated comments.}
    \label{fig: case}
    \vspace{-15pt}
\end{figure}

\paragraph{Case Study}
We study a specific case of news article and its LLM-generated comments in Figure \ref{fig: case}. The red area indicates that the generated comments match the user attributes about partisanship and age groups. The green areas indicate strong continuity in the comment chain. Overall, the example showcases the effectiveness of \ourmethod{} in generating diverse comments that ground news articles and facilitate characterization.

\section{Related Work}
Existing fake news detection methods \citep{zeng2022early, biamby2022twitter, DBLP:conf/acl/MendesCXR23, DBLP:conf/emnlp/SungBH23, xu2023leveraging, liao2023muser} mostly fall into text-based \citep{pelrine2021surprising, jin2022towards, chen2023causal} and graph-based approaches \citep{wu2022cross, zhou2022fake, karami2023silence, DBLP:conf/acl/FengTZLT23, lin2023zero, phan2023fake, chang2023beyond, ma2023kapalm}. 
Text-only approaches take news context and employ NLP methodologies for classification such as recurrent neural networks \citep{goonathilake2020cnn, DBLP:conf/acl/LiuWL23}, attention mechanism \citep{shu2019defend, dun2021kan}, and pre-trained language models \citep{hartl2022applying}. 
In addition to solely considering news content, graph-based approaches first construct networks composed of entities such as news articles, sources \citep{nguyen2020fang}, users \citep{shu2019role, dou2021user}, and more. These approaches then employ graph neural networks \citep{bian2020rumor, zhang2024bayesian} for classification. 
Among graph-based approaches, the widely used is to employ comments, \ie, user reactions to news article on social media \citep{yang2021rumor, tian2022duck, mehta2022tackling, DBLP:conf/emnlp/YangG0L023, russo2023countering, min2023pesto}. In this work, we seek to employ LLMs to generate synthetic comments from diverse perspectives to complement the scarce and incomplete comment networks in real-world datasets \citep{jung2020caution, micallef2020role, heidari2021bert}.

With the advent of autoregressive large language models, previous works have attempted to gauge their risks and generate misinformation with LLMs \citep{zellers2019defending, fung2021infosurgeon, DBLP:conf/acl/HuangMNCJ23, wang2023attacking}. They find that LLMs are capable of generating misinformation that is challenging to detect and characterize \citep{huang2022faking, chen2023can, DBLP:conf/emnlp/PanPCNKW23, goldstein2023generative, su2023fake, xu2023earth, uchendu2023does}. 
On the other hand, researchers have attempted to employ LLMs off-the-shelf for misinformation research through prompting and in-context learning \citep{stiff2022detecting, gabriel2022misinfo, kim2023covid, DBLP:conf/emnlp/PelrineITRGCGR23, russo2023countering, jiang2023disinformation, nakshatri2023using, sundriyal2023chaos, su2023adapting, li2023revisit, chen2023can2, feng2023cook, DBLP:conf/acl/YueZZSW23, yang2023rumor, choi2023automated}. 
We argue that LLMs face challenges of hallucination \citep{ji2023survey, du2023quantifying}, factuality \citep{kandpal2023large, DBLP:journals/tgdk/PanRKSCDJO0LBMB23}, and temporal knowledge update \citep{feng2023knowledge, luo2024chain}: as a result, they could not be directly used off-the-shelf for predicting a True-of-False label since they lack accurate and up-to-date information about real-world news events, while such information is crucial in characterizing fake news campaigns. 
To this end, we identify three key stages in evaluating news veracity and propose strategies to integrate LLMs in countering online misinformation campaigns.

\section{Conclusion}
We propose \ourmethod{} for identifying fake news where LLMs could be incorporated as part of the pipeline. First, we employ LLMs to generate news reactions from diverse perspectives and simulate user-news networks. Second, we design six explainable proxy tasks that help identify misinformation. LLMs perform these tasks and generate explanations to produce experts specializing in various aspects of news articles. Finally, we develop three strategies for LLMs to merge task-specific experts and provide an overall prediction. Extensive experiments demonstrate that \ourmethod{} achieves state-of-the-art performance on three tasks across seven datasets, presenting a misinformation detector better calibrated and better grounded in diverse perspectives.

\section*{Acknowledgement}
This work was supported by the National Nature Science Foundation of China (No. 62192781, No. 62272374), the Natural Science Foundation of Shaanxi Province (2024JC-JCQN-62), the National Nature Science Foundation of China (No. 62202367, No. 62250009, No. 62137002), Project of China Knowledge Center for Engineering Science and Technology, and Project of Chinese academy of engineering ``The Online and Offline Mixed Educational Service System for `The Belt and Road' Training in MOOC China''. We would like to express our gratitude for the support of K. C. Wong Education Foundation.
\section*{Limitation}
While \ourmethod{} could generate synthetic news reactions from diverse perspectives and form networks of user-news interactions, the iterative process with LLMs in computationally heavy. Scaling our solution to the real-world scale of millions of real-time news reactions could be challenging, while we expect efficient LLM inference approaches could help alleviate this limitation.

While we develop six proxy tasks for LLMs to generate explanations and enrich news contexts, they may not be able to fully tap into the diverse capabilities of LLMs and their potential for evaluating the veracity of news articles. Future work could focus on automatically generating and proposing proxy tasks for a more general LLM-as-enhancer framework.

\section*{Ethics Statement}
The development of fake news detectors is essential in countering online misinformation campaigns. This research demonstrates that LLMs could be integrated as part of the news analysis pipeline. However, it may increase the risk of dual-use, where malicious actors may develop advanced misinformation campaigns that are evasive to LLM-generated comments and explanations. We will establish controlled access to ensure that the data and trained model checkpoint are only publicly available to researchers.

LLMs have been widely shown to have inherent social biases \citep{DBLP:conf/fat/BenderGMS21, jin2021transferability, DBLP:conf/acl/Shaikh0HBY23}, and such biases could have an impact on fake news detection~\citep{DBLP:conf/acl/FengPLT23}. Informed by LLMs' internal biases, stereotypes, and spurious correlations, \ourmethod{} might struggle to simulate certain demographic groups and provide incorrect explanations of news articles. We argue that the predictions of \ourmethod{} should be interpreted as an initial screening, while content moderation decisions should be made with experts in the loop.

\bibliography{custom}

\newpage

\appendix

\section{Methodology Details}
\subsection{User Attribute Details}
\label{app: user_attribute}
We simulate each synthetic user as an intersection of seven categories, and the detailed attribute descriptions of each category are as follows: 
\begin{itemize}[leftmargin=*]
    \item \textbf{gender}: ``\textit{You are male.}''; ``\textit{You are female.}''
    \item \textbf{age}: ``\textit{You are under 17 years old.}''; ``\textit{You are 18 to 29 years old.}''; ``\textit{You are 30 to 49 years old.}''; ``\textit{You are 50 to 64 years old.}''; ``\textit{You are over 65 years old.}''
    \item \textbf{ethnicity}: ``\textit{Racially, you are White.}''; ``\textit{Racially, you are Black.}''; ``\textit{Racially, you are Hispanic.}''
    \item \textbf{education level}: ``\textit{Educationally, you are a college grad.}''; ``\textit{Educationally, you haven't graduated from college.}''; ``\textit{Educationally, you have a high school diploma or less.}''
    \item \textbf{family income}: ``\textit{Financially, your annual family income is more than 75,000.}''; ``\textit{Financially, your annual family income is 30,000 to 74,999.}''; ``\textit{Financially, your annual family income is less than 30,000.}''
    \item \textbf{political leaning}: ``\textit{Politically, you are a Republican.}''; ``\textit{Politically, you are a Democrat.}''
    \item \textbf{voter registration}: ``\textit{Meanwhile, you are registered to vote.}''; ``\textit{Meanwhile, you are probably registered to vote.}''; ``\textit{Meanwhile, you are not registered to vote.}''
\end{itemize}

\begin{table}[]
    \centering
    \begin{tabularx}{\linewidth}{m{0.95\linewidth}}
    \toprule[1pt]
        You are a social media user. You are female. You are 18 to 29 years old. Racially, you are Hispanic. Financially, your annual family income is 30,000 to 74,999. Educationally, you are a college grad. Politically, you are a Republican. Meanwhile, you are probably registered to vote.\\
    \bottomrule[1pt]
    \end{tabularx}
    \caption{An Example of a synthetic social media user prompt.}
    \label{tab: user_profile}
\end{table}
We uniformly sample each value for the seven attributes to represent a social media user. We then concatenate these attributes prefixed with ``\textit{You are a social media user.}'' as the prompt for the synthetic user. Table \ref{tab: user_profile} illustrates a complete example of a synthetic social media user prompt.

\subsection{User-News Networks Details}
\label{app: user-news_networks}
\begin{algorithm}[t]
    \caption{Pseudo-code of user-news network generation.}
    \label{algo: networks}
    \KwIn{news content $\boldsymbol{s}$; graph size $m$; $\alpha$ to control the probability of commenting on the news; $\beta$ to control the balance of tree height and width; $k$ to control candidate set size}
    \KwOut{user-news network $\mathcal{G}(\mathcal{V}, \mathcal{E})$}
    $\mathcal{V}$ = [$\boldsymbol{s}$]\\
    $\mathcal{E}$ = []\\
    $\mathcal{F}$ = [$\boldsymbol{s}$]\\
    $\mathcal{H}$ denoting height of each node\\
    $\mathcal{W}$ denoting width of each node\\
    \While{$\|V\|\leq m$}{
        $\boldsymbol{u}$ = $\psi()$\\
        $p \sim U(0, 1)$\\
        \eIf{$p <= \alpha$}{
            \textsc{prompt} = $\phi_1(\boldsymbol{s}, \boldsymbol{u})$\\
            $\boldsymbol{f}$ = $\boldsymbol{s}$\\
        }{
            $\mathcal{P}$ = $\beta * \mathcal{H}+(1-\beta)*\mathcal{W}$\\
            $\{\boldsymbol{c}^i\}_i$ = Sample($\mathcal{V}, \mathcal{P}, k$)\\
            \textsc{select} = $\phi_3(\boldsymbol{s}, \boldsymbol{u}, \{\boldsymbol{C}^i\}_i)$\\
            $j$ = LLM(\textsc{select})\\
            $\boldsymbol{f}$ = $\boldsymbol{C}$ = $\boldsymbol{C}^j$\\
            \textsc{prompt} = $\phi_2(\boldsymbol{s}, \boldsymbol{u}, \boldsymbol{C})$\\
        }
        $\boldsymbol{s}_{\textit{out}}$ = LLM(\textsc{prompt})\\
        $\mathcal{V}$.append($\boldsymbol{s}_{\textit{out}}$)\\
        $\mathcal{E}$.append($(\boldsymbol{s}_{\textit{out}}, \boldsymbol{f})$)\\
        $\mathcal{F}$.append($\boldsymbol{f}$)\\
        update $\mathcal{H}$ and $\mathcal{W}$\\
    }
    \KwReturn{$\mathcal{G}(\mathcal{V}, \mathcal{E})$}
\end{algorithm}
Our generated user-news interaction networks $\mathcal{G}$ forms a tree structure, where $\|\mathcal{V}\|=\|\mathcal{E}\|+1$. 
To control the iterative process of generating user-news networks, we design hyperparameters $\alpha$ to control the probability of commenting on news and $\beta$ to control the balance of tree height and width. We present an algorithmic summary of the generation process in Algorithm \ref{algo: networks}, where $\phi_1(\boldsymbol{s}, \boldsymbol{u})$, $\phi_2(\boldsymbol{s}, \boldsymbol{u}, \boldsymbol{C})$, and $\phi_3(\boldsymbol{s}, \boldsymbol{u}, \{\boldsymbol{C}^i\}_i)$ denotes \textit{comment on news}, \textit{Comment on a comment}, and \textit{Select a comment to comment} prompt generation process, $\psi()$ denotes the diverse user attribute prompt generation process as mentioned in Appendix \ref{app: user_attribute}, $\mathrm{Sample}(\mathcal{V}, \mathcal{P}, k)$ denote the sample function that sample $k$ instances from $\mathcal{V}$ according to the probability $\mathcal{P}$. Table \ref{tab: network_prompt} presents the prompt templates.

\subsection{Explainable Proxy Task Details}
\label{app: proxy}
We propose four proxy tasks to enhance news articles, the details of \textbf{Sentiment Analysis}, \textbf{Framing Detection}, and \textbf{Propaganda Tactics Detection} are as follows:
\begin{itemize}[leftmargin=*]
    \item \textbf{Sentiment Analysis}: we employ six basic emotions: \textit{anger}; \textit{disgust}; \textit{fear}; \textit{happiness}; \textit{sadness}; and \textit{surprise}.
    \item \textbf{Framing Detection}: we employ 14 news frames: \textit{Economic}; \textit{Capacity and resources}; \textit{Morality}; \textit{Fairness and equality}; \textit{Legality, constitutionality and jurisprudence}; \textit{Policy prescription and evaluation}; \textit{Crime and punishment}; \textit{Security and defense}; \textit{Health and safety}; \textit{Quality of life}; \textit{Cultural identity}; \textit{Among public opinion}; \textit{Political}; \textit{External regulation and reputation}.
    \item \textbf{Propaganda Tactics Detection}: we employ 19 propaganda tactics: \textit{Conversation Killer}; \textit{Whataboutism}; \textit{Doubt}; \textit{Straw Man}; \textit{Red Herring}; \textit{Loaded Language}; \textit{Appeal to Fear-Prejudice}; \textit{Guilt by Association}; \textit{Flag Waving}; \textit{False Dilemma-No Choice}; \textit{Repetition}; \textit{Appeal to Popularity}; \textit{Appeal to Authority}; \textit{Name Calling-Labeling}; \textit{Slogans}; \textit{Appeal to Hypocrisy}; \textit{Exaggeration-Minimisation}; \textit{Obfuscation-Vagueness-Confusion}; \textit{Causal Oversimplification}. 
\end{itemize}
Table \ref{tab: proxy_prompt} presents the presents the prompt template of each proxy task.

\subsection{LLM-Based Expert Ensemble Details}
\label{app: ensemble}
We propose three LLM-based approaches to selectively integrate the prediction of each expert. The description $\boldsymbol{d}_i$ of each expert $e_i$ is as follows:
\begin{itemize}[leftmargin=*]
    \item \textbf{w/o expert}: \textit{This expert is comprehensive}. 
    \item \textbf{sentiment}: \textit{This expert focuses on the emotion of this news}.
    \item \textbf{framing}: \textit{This expert focuses on the framing of this news}.
    \item \textbf{propaganda tactics}: \textit{This expert focuses on the propaganda tactics of this news.}.
    \item \textbf{retrieval}: \textit{This expert focuses on the external knowledge of this news}.
    \item \textbf{stance}: \textit{This expert focuses on the stance of related comments on this news}.
    \item \textbf{relation}: \textit{This expert focuses on the relation of related comments on this news}.
\end{itemize}
to obtain the confidence score, we employ a softmax operator $\boldsymbol{score}$ for binary classification and absolute value operator for multi-label classification. 
We provide the prompt templates in Table \ref{tab: ensemble_prompt}.

\section{Experiment Setting Details}
\subsection{Dataset Details}
\label{app: datasets}
We evaluate \ourmethod{} and baselines on three tasks related to fake news detecton.

1) Fake News Detection:
\begin{itemize}[leftmargin=*]
    \item \textbf{Pheme} \citep{buntain2017automatically} is a dataset of potential rumors on Twitter and journalistic assessments of their accuracies.
    \item \textbf{LLM-mis} \citep{chen2023can} is a LLM-generated misinformation dataset with different LLM generators and generation approaches.
\end{itemize}
2) Framing Detection:
\begin{itemize}[leftmargin=*]
    \item \textbf{MFC} \citep{card.2015} contains labeled and unlabeled articles on six issues from 14 newspapers covering the years 1990-2014, though some issues have broader coverage. The issues include climate; the death penalty; gun control; immigration; same-sex sex; and tobacco. We sample the labeled articles as a benchmark. 
    \item \textbf{SemEval-23F} \citep{piskorski2023semeval} aim to identify one or more frames used in an article from a pool of 14 generic frames: Security and defense; Fairness and equality; Political; Capacity and resources; Economic; Morality; Policy prescription and evaluation; Legality Constitutionality and jurisprudence; External regulation and reputation; Quality of life; Health and safety; Cultural identity; Crime and punishment; and Public opinion.
\end{itemize}
3) Propaganda Tactic Detection:
\begin{itemize}[leftmargin=*]
    \item \textbf{Generated} is a benchmark generated by \textit{ChatGPT}. We first determine 4 topics: Russia and Ukraine; Palestine and Israel; the Republican Party; and the Democratic Party. Around these topics, we generated 5 paragraphs for each tactic.
    \item \textbf{SemEval-20} \citep{DBLP:data/10/MartinoBWPN20} contains 14 possible propaganda tactics: Appeal to fear-prejudice; Black-and-White Fallacy; Name Calling, Labeling; Slogans; Whataboutism,Straw Men,Red Herring; Exaggeration, Minimisation; Loaded Language; Repetition; Causal Oversimplification; Bandwagon,Reductio ad hitlerum; Flag-Waving; Thought-terminating Cliches; Appeal to Authority; and Doubt. This benchmark merges some tactics into one category.
    \item \textbf{SemEval-23F} \citep{piskorski2023semeval} contains 6 main categories: Attack on reputation; Justification; Simplification; Distraction; Call; and Manipulative wording. It contains 19 propaganda tactics: Conversation Killer; False Dilemma-No Choice; Appeal to Popularity; Doubt; Flag Waving; Slogans; Whataboutism; Straw Man; Loaded Language; Name Calling-Labeling; Obfuscation-Vagueness-Confusion; Appeal to Fear-Prejudice; Causal Oversimplification; Red Herring; Repetition; Exaggeration-Minimisation; Appeal to Authority; Guilt by Association; and Appeal to Hypocrisy.
\end{itemize}

We randomly sample 1,000 instances from each benchmark (select all if there are less than 1,000 instances), and divided the training set, validation set, and test set according to the ratio of 7:2:1.
\subsection{Bseline Details}
\label{app: baseline}
\begin{itemize}[leftmargin=*]
    \item \textsc{zero-shot} asks LLMs to conduct detection.
    \item \textsc{few-shot} first provides LLMs with some pairs of news instances and labels and then asks LLMs to conduct detection.
    \item \textsc{retrieval}-augmented generation first provides LLMs with the external knowledge retrieved from Wikipedia, which is the same as \textbf{Knowledge Retrieval} proxy task. It then asks LLM to conduct detection.
    \item \textsc{F3} \textit{Z-CoT} \citep{lucas2023fighting} uniquely leverages LLMs’ self-formulated rationales by integrating a standard instruction with the simple phrase, “Let’s think step by step known as Chain of Thoughts (CoT).”
    \item \textsc{F3} \textit{DeF-Gen} \citep{lucas2023fighting} focuses contextually, emphasizing deductive and abductive reasoning.
    \item \textsc{TAPE w/o graph} \citep{he2023harnessing}  focus on leveraging LLMs to capture textual information as features, which can subsequently enhance GNN performance on downstream tasks. Here we only employ the textual information generated by LLMs to enhance news content.
    \textsc{DeBERTa} \citep{deberta} leverages the pre-trained language model DeBERTa to encode news content, then feed them into an MLP classifier.
    \item \textsc{k-hops} \citep{huang2023can} incorporates randomly selected neighbors into the prompt, where the idea behind is to aggregate information from neighboring nodes, following GCN.
    \item \textsc{k-attention} \citep{huang2023can} is designed to weigh the influence of neighboring nodes during the prediction process, following GAT.
    \item \textsc{TAPE w/ graph} \citep{he2023harnessing} puts the enhanced news content into the user-news network and employs graph neural networks to conduct detection.
    \item \textsc{GCN} \citep{DBLP:conf/iclr/KipfW17} adopt multiple GNN layers and a meaning pooling to obtain the user-news network representations. 
    \item \textsc{dEFEND} \citep{shu2019defend} conducts explainable detection by the attention weights, we set maximum sentence length and maximum comment length as 96, maximum sentence count as 64, and maximum comment count as 10 to reproduce so that the approach is applicable to our tasks and datasets.
    \item \textsc{RvNN} \citep{ma2018rumor} proposes two recursive neural model stratages: bottom-up and top-down tree-structured neural networks. We employ the top-down structure.
    \item \textsc{Hypehn} \citep{grover2022public} is a discourse-aware hyperbolic spectral co-attention network. It is a fusion of hyperbolic graph representation learning with a novel Fourier co-attention mechanism in an attempt.
    \item \textsc{GET} \citep{xu2022evidence} models claims and related evidences as graph-structured data and capture the long-distance semantic dependency among dispersed relevant snippets via neighborhood propagation.
    \item \textsc{WSDMS} \citep{DBLP:conf/emnlp/YangG0L023} needs bag-level labels for training but possesses the capability to infer both sentence-level misinformation and article-level veracity, facilitated by pertinent social media conversations meticulously contextualized with news sentences.
\end{itemize}
For the LLM-based baseline, we provide the prompt templates in Table \ref{tab: LLM_baseline_prompts}. Each baseline prompt template contains a task-related prompt to describe the task and a baseline-related prompt.

\begin{table}[]
    \centering
    \begin{tabular}{c|c}
    \toprule[1.5pt]
         \textbf{Hyperparameter}&  \textbf{Value}\\
         \midrule[1pt]
         optimizer&Adam\\
         learning rate&10$^{-4}$\\
         weight decay&10$^{-5}$\\
         dropout&0.5\\
         hidden dim&1024\\
         embedding dim&1024\\
         GNN layers&2\\
         maximum epochs&100\\
         temperature for reaction generation $\tau$&0.6\\
         temperature&0.1\\
         \bottomrule[1.5pt]
    \end{tabular}
    \caption{Hyperparameter settings of \ourmethod{}.}
    \label{tab: hyper}
\end{table}

\subsection{Hyperparameters}
The hyperparameter settings of \ourmethod{} are presented in Table \ref{tab: hyper} to facilitate reproduction.

\section{Additional Results}
\label{app: exp}
We leverage \textit{Mistral-7B} \citep{jiang2023mistral}, \textit{LLaMA2-70B} \citep{touvron2023llama}, and \textit{ChatGPT} as the base LLMs. Tabel \ref{tab: more_res} presents more results. \ourmethod{} still outperforms other baselines. 

For context, \ourmethod{} has three components: Diverse Reaction Generation; Explainable Proxy Tasks; and LLM-Based Expert Ensemble. On the component level, we conduct more ablation studies as follows:
\begin{itemize}
    \item w/o Diverse Reaction Generation: we remove the network part and only employ the news content and related proxy task outputs.
    \item w/o Explainable Proxy Tasks: we remove all proxy tasks and conduct experiments with news content and generated content.
    \item w/o LLM-Based Expert Ensemble: we replace LLM-based ensembling with simple majority voting: majority vote; confidence weighted sum; and train the weights on validation set.
\end{itemize}

We present the results of the ablation study in Table \ref{tab: ablation}. Every module of \ourmethod{} could improve the fake news detection performance.
\section{Additional Analysis}
\label{app: analysis}
\subsection{Model Robustness to Comments (cont.)}
\begin{figure*}
    \centering
    \includegraphics[width=\linewidth]{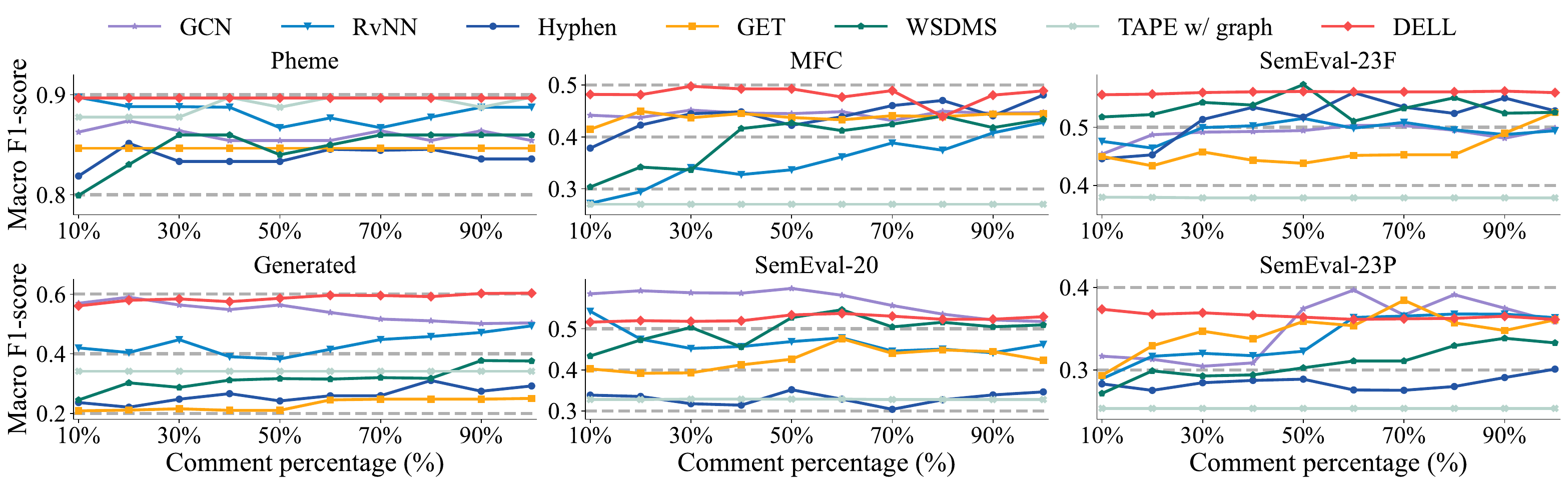}
    \caption{Performance of \ourmethod{} and baselines when the comments are gradually removed. \ourmethod{} shows great robustness to the availability of comments.}
    \label{fig: comment_drop_appendix}
\end{figure*}

Figure \ref{fig: comment_drop_appendix} presents the results of the other six benchmarks. On other benchmarks, \ourmethod{} drops the least in performance with reduced comments. Specifically, \ourmethod{} almost keeps the same on \textbf{Pheme} and drops 11.8\% on \textbf{MFC}, 1.1\% on \textbf{SemEval-23F}, 7.1\% on \textbf{Generated}, 3.9\% on \textbf{SemEval-20}, and 3.3\% on \textbf{SemEval-23P}.
\subsection{Expert Selection (cont.)}
\begin{figure*}
    \centering
    \includegraphics[width=\linewidth]{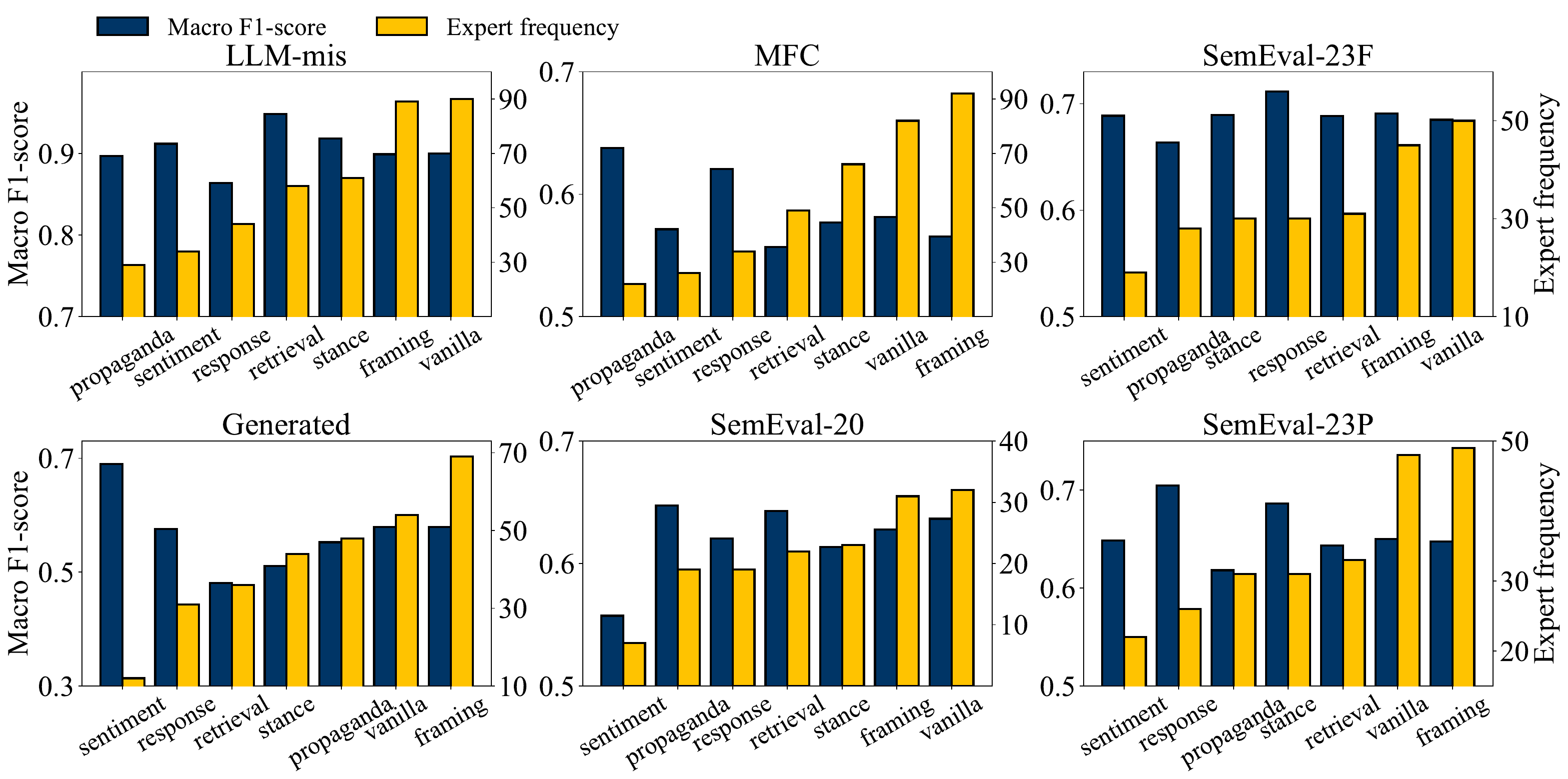}
    \caption{The frequency of expert selections and performance when a particular expert is selected in the \textit{Selective} approach.}
    \label{fig: expert_ana_appendix}
\end{figure*}

Figure \ref{fig: expert_ana_appendix} presents the results of the other six benchmarks. Besides this experiment, we also examine the count of experts in every selection and corresponding performance. The results are shown in Table \ref{tab: expert_cnt}.

\begin{table*}[]
    \centering
    \resizebox{\textwidth}{!}{
    \begin{tabular}{l|cc|cc|cc|cc|cc|cc|cc}
    \toprule[1.5pt]
        \multirow{3}{*}{\textbf{Method}}&\multicolumn{4}{c|}{\textbf{Fake News Detection}}&\multicolumn{4}{c|}{\textbf{Framing Detection}}&\multicolumn{6}{c}{\textbf{Propaganda Tactic Detection}}\\
        &\multicolumn{2}{c|}{\textbf{Pheme}}&\multicolumn{2}{c|}{\textbf{LLM-mis}}&\multicolumn{2}{c|}{\textbf{MFC}}&\multicolumn{2}{c|}{\textbf{SemEval-23F}}&\multicolumn{2}{c|}{\textbf{Generated}}&\multicolumn{2}{c|}{\textbf{SemEval-20}}&\multicolumn{2}{c}{\textbf{SemEval-23P}}\\
        &MaF&MiF&MaF&MiF&MaF&MiF&MaF&MiF&MaF&MiF&MaF&MiF&MaF&MiF\\
    \midrule[1pt]
        \textbf{\ourmethod{}}&.820&.820&.928&.930&.509&.603&.572&.718&.598&.577&.525&.636&.386&.643\\
        \midrule[1pt]
        w/o Diverse Reaction \textit{Single}&.790&.790&.896&.900&.433&.575&.528&.663&.551&.552&.516&.602&.407&.604\\
        w/o Diverse Reaction \textit{Vanilla}&.800&.800&.907&.910&.440&.581&.521&.673&.522&.519&.490&.599&.370&.619\\
        w/o Diverse Reaction \textit{Confidence}&.789&.790&.875&.880&.429&.581&.361&.544&.566&.588&.524&.613&.376&.634\\
        w/o Diverse Reaction \textit{Selective}&.810&.810&.887&.890&.477&.594&.521&.670&.528&.537&.542&.629&.365&.606\\
        \midrule[1pt]
        w/o Explainable Proxy Tasks&.790&.790&.915&.920&.417&.577&.518&.704&.543&.556&.504&.596&.364&.620\\
        \midrule[1pt]
        \textit{Majority Vote}&.830&.830&.917&.920&.418&.576&.555&.703&.580&.593&.544&.647&.377&.644\\
        \textit{Confidence weight}&.820&.820&.917&.920&.458&.593&.583&.705&.578&.550&.504&.613&.369&.661\\
        \textit{Train on Validation Set}&.800&.800&.897&.900&.496&.585&.579&.674&.566&.527&.546&.622&.407&.638\\
        
    \bottomrule[1.5pt]
    \end{tabular}
    }
    \vspace{-5pt}
    \caption{Performance of variants of \ourmethod{}. The ablation study results illustrate that every module of \ourmethod{} is helpful for fake news detection.}
    \label{tab: ablation}
    \vspace{-15pt}
\end{table*}

\begin{table*}[]
    \centering
    \resizebox{\textwidth}{!}{
    \begin{tabular}{l|cc|cc|cc|cc|cc|cc|cc}
    \toprule[1.5pt]
        \multirow{3}{*}{\textbf{Method}}&\multicolumn{4}{c|}{\textbf{Fake News Detection}}&\multicolumn{4}{c|}{\textbf{Framing Detection}}&\multicolumn{6}{c}{\textbf{Propaganda Tactic Detection}}\\
        &\multicolumn{2}{c|}{\textbf{Pheme}}&\multicolumn{2}{c|}{\textbf{LLM-mis}}&\multicolumn{2}{c|}{\textbf{MFC}}&\multicolumn{2}{c|}{\textbf{SemEval-23F}}&\multicolumn{2}{c|}{\textbf{Generated}}&\multicolumn{2}{c|}{\textbf{SemEval-20}}&\multicolumn{2}{c}{\textbf{SemEval-23P}}\\
        &MaF&MiF&MaF&MiF&MaF&MiF&MaF&MiF&MaF&MiF&MaF&MiF&MaF&MiF\\
    \midrule[1pt]
        \multicolumn{15}{l}{LLM-based baselines with \textit{Mistral-7B}.}\\
        \textsc{zero-shot}     &.450&.450&.517&.560&.247&.265&.371&.431&.189&.202&.293&.408&.207&.274\\
        \textsc{few-shot}      &.385&.390&.639&.660&.259&.303&.376&.432&.170&.217&.382&.466&.306&.372\\
        \textsc{retrieval}     &.466&.480&.648&.670&.216&.260&.285&.383&.116&.134&.280&.389&.234&.310\\
        \textsc{TAPE w/o graph}&.789&.790&.895&.900&.278&.497&.409&.610&.340&.353&.320&.595&.259&.632\\
        \textsc{k-hops}        &.301&.430&.533&.550&.255&.273&.377&.489&.107&.110&.286&.406&.156&.243\\
        \textsc{k-attention}   &.310&.420&.483&.510&.280&.336&.414&.508&.143&.145&.433&.474&.264&.312\\
        \textsc{TAPE w graph}  &.800&.800&.897&.900&.270&.485&.379&.633&.341&.358&.328&.598&.253&.608\\
    \midrule[1pt]
        \multicolumn{15}{l}{LLM-based baselines with \textit{LLaMA2-70B}.}\\
        \textsc{zero-shot}     &.403&.410&.650&.650&.331&.374&.380&.493&.176&.178&.143&.228&.044&.140\\
        \textsc{few-shot}      &.322&.420&.670&.670&.312&.357&.396&.480&.117&.119&.404&.452&.335&.415\\
        \textsc{retrieval}     &.513&.520&.672&.680&.315&.354&.325&.483&.163&.167&.109&.186&.073&.150\\
        \textsc{TAPE w/o graph}&.748&.750&.856&.860&.376&.581&.448&.654&.381&.427&.358&.613&.244&.612\\
        \textsc{k-hops}        &.310&.420&.634&.690&.327&.386&.394&.521&.204&.180&.189&.292&.054&.167\\
        \textsc{k-attention}   &.400&.410&.750&.760&.306&.378&.414&.539&.198&.196&.312&.429&.109&.203\\
        \textsc{TAPE w graph}  &.752&.760&.857&.860&.392&.575&.453&.670&.370&.420&.320&.592&.244&.632\\
    \midrule[1pt]
    \midrule[1pt]
    \multicolumn{15}{l}{All expert performance of \ourmethod{} with \textit{ChatGPT}.}\\
        \textsc{vanilla}   &.790&.790&.915&.920&.417&.577&.518&.704&.543&.556&.504&.596&.364&.620\\
        \textsc{sentiment} &.780&.780&.867&.870&.413&.552&.536&.684&.510&.492&.499&.578&.343&.650\\
        \textsc{framing}   &.810&.810&.887&.890&.446&.571&.509&.658&.509&.541&.520&.613&.375&.618\\
        \textsc{propaganda}&.780&.780&.858&.860&.458&.598&.487&.604&.506&.525&.496&.583&.363&.606\\
        \textsc{retrieval} &.779&.780&.897&.900&.450&.570&.512&.646&.522&.520&.513&.589&.370&.601\\
        \textsc{stance}    &.780&.780&.917&.920&.435&.571&.532&.683&.517&.547&.496&.606&.376&.631\\
        \textsc{response}  &.780&.780&.928&.930&.428&.582&.506&.695&.536&.538&.493&.618&.364&.646\\
     \midrule[1pt]
        \multicolumn{15}{l}{expert ensemble of \ourmethod{} with \textit{Mistral-7B}.}\\
        \textsc{\ourmethod{} \textit{Vanilla}}   &.770&.770&.888&.890&.411&.526&.577&.689&.551&.519&.513&.618&.337&.566\\
        \textsc{\ourmethod{} \textit{Confidence}}&.789&.790&.866&.870&.458&.571&.539&.676&.539&.507&.484&.590&.347&.623\\
        \textsc{\ourmethod{} \textit{Selective}} &.820&.820&.917&.920&.478&.579&.570&.700&.608&.577&.493&.608&.367&.662\\
     \midrule[1pt]
        \multicolumn{15}{l}{expert ensemble of \ourmethod{} with \textit{LLaMA2-70B}.}\\
        \textsc{\ourmethod{} \textit{Vanilla}}   &.722&.730&.906&.910&.453&.582&.549&.700&.579&.553&.563&.655&.382&.646\\
        \textsc{\ourmethod{} \textit{Confidence}}&.624&.670&.894&.900&.421&.569&.509&.685&.555&.549&.541&.649&.371&.632\\
        \textsc{\ourmethod{} \textit{Selective}} &.810&.810&.897&.900&.457&.592&.573&.704&.575&.547&.505&.615&.367&.655\\
    \bottomrule[1.5pt]
    \end{tabular}
    }
    \caption{Performance of \ourmethod{} and baselines using other LLMs on seven datasets from three tasks related to fake news detection. \ourmethod{} still outperforms other baselines.}
    \label{tab: more_res}
\end{table*}

\begin{table*}[]
    \centering
    \begin{tabular}{lc|cccccccc}
        \toprule[1.5pt]
        \multicolumn{2}{c|}{Benchmark}&0&1&2&3&4&5&6&7\\
        \midrule[1pt]
        \multirow{2}{*}{Pheme}&instance count&0&1&5&49&22&19&4&0\\\cline{2-10}
        &macro f1-score&nan&1.00&1.00&.816&.818&.789&.750&nan\\
        \midrule[1pt]
        \multirow{2}{*}{LLM-mis}&instance count&0&0&2&37&32&16&9&4\\\cline{2-10}
        &macro f1-score&nan&nan&.500&.946&.813&1.00&.889&1.00\\
        \midrule[1pt]
        \multirow{2}{*}{MFC}&instance count&0&0&3&43&36&17&0&1\\\cline{2-10}
        &macro f1-score&nan&nan&.571&.602&.548&.583&nan&.800\\
        \midrule[1pt]
        \multirow{2}{*}{SemEval-23F}&instance count&0&0&2&11&15&12&8&4\\\cline{2-10}
        &macro f1-score&nan&nan&63.2&62.1&70.5&73.3&66.7&68.8\\
        \midrule[1pt]
        \multirow{2}{*}{Generated}&instance count&0&0&1&27&23&19&4&0\\\cline{2-10}
        &macro f1-score&nan&nan&1.00&.704&.500&.476&.600&nan\\
        \midrule[1pt]
        \multirow{2}{*}{SemEval-20}&instance count&0&1&2&8&15&8&4&0\\\cline{2-10}
        &macro f1-score&nan&0.833&0.800&0.615&0.604&0.615&0.667&nan\\
        \midrule[1pt]
        \multirow{2}{*}{SemEval-23P}&instance count&0&1&2&8&19&13&7&4\\\cline{2-10}
        &macro f1-score&nan&1.00&.625&.582&.682&.607&.644&.745\\
        \midrule[1pt]
    \end{tabular}
    \caption{The count of experts in every selection and corresponding performance in the \textit{selective} approach.}
    \label{tab: expert_cnt}
\end{table*}

\subsection{Case Study (cont.)}
Table \ref{tab: more_case_1}, \ref{tab: more_case_2}, and \ref{tab: more_case_3} provide more cases of the explanations of proxy tasks generated by \ourmethod{}. It illustrates that LLMs could generate reasonable explanations of proxy tasks, providing more information in identifying fake news.

\begin{table*}
    \centering
    \begin{tabularx}{\linewidth}{m{0.3\linewidth}|m{0.65\linewidth}}
        \toprule[1.5pt]
         \textbf{Comment on news}& 
         $\boldsymbol{u}$\newline
         You view a piece of news with the following content.\newline
         News: $\boldsymbol{s}$\newline
         Task: Please comment on this news on social media. Your comment is limited to 40 words.\newline
         Your comment: 
         \\
         \midrule[1pt]
         \textbf{Comment on a comment}&
         $\boldsymbol{u}$\newline
         You view a piece of news and a related comment chain on social media, and their contents are as follows.\newline
         News: $\boldsymbol{s}$\newline
         Comment 1: $\boldsymbol{c}_1$\newline
         Comment 2: $\boldsymbol{c}_2$\newline
         $\dots$\newline
         Comment $k$: $\boldsymbol{c}_k$\newline
         Task: Please reply to the last comment(comment $k$) on social media. Your reply is limited to 40 words.\newline
         Your reply: 
         \\
         \midrule[1pt]
         \textbf{Select a comment to comment}&
         $\boldsymbol{u}$\newline
         You view a piece of news and related comment chains on social media, and their contents are as follows.\newline
         News: $\boldsymbol{s}$\newline
         Comment Chain 1:\newline
         Comment 1: $\boldsymbol{c}_1^1$\newline
         Comment 2: $\boldsymbol{c}_2^1$\newline
         $\dots$\newline
         Comment $k_1$: $\boldsymbol{c}_{k_1}^1$\newline
         Comment Chain 2:\newline
         $\dots$\newline
         Comment Chain $n$:\newline
         $\dots$\newline
         Comment $k_n$: $\boldsymbol{c}_{k_n}^n$\newline
         Task: Please select a comment chain that you would most like to comment on. Answer the selected number and explain the reason.\newline
         Answer: 
         \\
         \bottomrule[1.5pt]
    \end{tabularx}
    \caption{Prompt templates of generating user-news networks}
    \label{tab: network_prompt}
\end{table*}

\begin{table*}
    \centering
    \begin{tabularx}{\linewidth}{m{0.3\linewidth}|m{0.65\linewidth}}
        \toprule[1.5pt]
         \textbf{Sentiment Analysis}& 
         News: $\boldsymbol{s}$\newline
         Task: Which emotions does the news contain? Please choose the three most likely ones: anger, disgust, fear, happiness, sadness, and surprise. Please provide your reasoning.\newline
         Answer:
         \\
         \midrule[1pt]
         \textbf{Framing Detection}& 
         News: $\boldsymbol{s}$\newline
         Task: Framing is a strategic device and a central concept in political communication for representing different salient aspects and perspectives to convey the latent meaning of an issue. Which framings does the news contain? Please choose the five most likely ones: Economic; Capacity and resources; Morality; Fairness and equality; Legality, constitutionality and jurisprudence; Policy prescription and evaluation; Crime and punishment; Security and defense; Health and safety; Quality of life; Cultural identity; Among public opinion; Political; External regulation and reputation. Please provide your reasoning.\newline
         Answer:
         \\
         \midrule[1pt]
         \textbf{Propaganda Tactics Detection}& 
         News: $\boldsymbol{s}$\newline
         Task:  Propaganda Tactics are methods used in propaganda to convince an audience to believe what the propagandist wants them to believe. Which propaganda techniques does the news contain? Please choose the five most likely ones: Conversation Killer; Whataboutism; Doubt; Straw Man; Red Herring; Loaded Language; Appeal to Fear-Prejudice; Guilt by Association; Flag Waving; False Dilemma-No Choice; Repetition; Appeal to Popularity; Appeal to Authority; Name Calling-Labeling; Slogans; Appeal to Hypocrisy; Exaggeration-Minimisation; Obfuscation-Vagueness-Confusion; Causal Oversimplification. Please provide your reasoning.\newline
         Answer:
         \\
         \midrule[1pt]
         \textbf{Knowledge Retrieval}& 
         News: $\boldsymbol{s}$\newline
         Task: Identify five named entities within the news above that necessitate elucidation for the populace to understand the news comprehensively. Ensure a diverse selection of the entities. The answer should in the form of python list.\newline
         Answer:
         \\
         \midrule[1pt]
         \textbf{Stance Detection}& 
         Task: Determine the stance of sentence 2 on sentence 1. Is it supportive, neutral or opposed? Provide your reasoning.\newline
         Sentence 1: $\boldsymbol{s}_1$\newline
         Sentence 2: $\boldsymbol{s}_2$\newline
         Answer:
         \\
         \midrule[1pt]
         \textbf{Response Characterization}& 
         Sentence 1: $\boldsymbol{s}_1$\newline
         Sentence 2: $\boldsymbol{s}_2$\newline
         Task: Sentence 1 and Sentence 2 are two posts on social networks. Please judge whether the sentence 2 replies to the sentence 1. Answer yes or no and provide the reasoning.\newline
         Answer:
         \\
         \bottomrule[1.5pt]
    \end{tabularx}
    \caption{Prompt templates of each proxy task.}
    \label{tab: proxy_prompt}
\end{table*}

\begin{table*}
    \centering
    \begin{tabularx}{\linewidth}{m{0.15\linewidth}|m{0.8\linewidth}}
        \toprule[1.5pt]
         \textbf{Vanilla}& 
         News: $\boldsymbol{s}$\newline
         Some experts give predictions about the news.\newline
         Expert 1: $\boldsymbol{d}_1$. The expert predicts the label of this news is $\boldsymbol{\ell}_1$.\newline
         Expert 2: $\boldsymbol{d}_2$. The expert predicts the label of this news is $\boldsymbol{\ell}_2$.\newline
         $\dots$\newline
         Expert 7: $\boldsymbol{d}_7$. The expert predicts the label of this news is $\boldsymbol{\ell}_7$.\newline
         Question: Based on the analysis of experts, please judge the final label of this news. Give the label in the form of ``[your answer]'', do not give any explanation.\newline
         Label:
         \\
         \midrule[1pt]
         \textbf{Confidence}& 
         News: $\boldsymbol{s}$\newline
         Some experts give predictions about the news.\newline
         Expert 1: $\boldsymbol{d}_1$. The expert predicts the label of this news is $\boldsymbol{\ell}_1$. The confidence scores are $\boldsymbol{score}_1$.\newline
         Expert 2: $\boldsymbol{d}_2$. The expert predicts the label of this news is $\boldsymbol{\ell}_2$. The confidence scores are $\boldsymbol{score}_2$.\newline
         $\dots$\newline
         Expert 7: $\boldsymbol{d}_7$. The expert predicts the label of this news is $\boldsymbol{\ell}_7$. The confidence scores are $\boldsymbol{score}_7$.\newline
         Question: Based on the analysis of experts, please judge the final label of this news. Give the label in the form of ``[your answer]'', do not give any explanation.\newline
         Label:
         \\
         \midrule[1pt]
         \textbf{Selective}&
          News: $\boldsymbol{s}$\newline
          Expert 1: $\boldsymbol{d}_1$.\newline
          Expert 2: $\boldsymbol{d}_2$.\newline
         $\dots$\newline
          Expert 7: $\boldsymbol{d}_7$.\newline
          To understand this news, which expert knowledge do you need? Return a Python list, e.g. [expert 1, expert 2, expert 6].
          \\
         \bottomrule[1.5pt]
    \end{tabularx}
    \caption{Prompt templates of our proposed LLM-based ensemble approaches.}
    \label{tab: ensemble_prompt}
\end{table*}

\begin{table*}
    \centering
    \begin{tabularx}{\linewidth}{m{0.3\linewidth}|m{0.65\linewidth}}
        \toprule[1.5pt]
        \textbf{Fake News Detection}&
        Task: Please determine whether the news is real or fake.
        \\
        \midrule[1pt]
        \textbf{Framing Detection}&
        Task: Framing is a strategic device and a central concept in political communication for representing different salient aspects and perspectives to convey the latent meaning of an issue. Which framings does the news contain? Please choose from: {\textit{the candidate label set of a specific dataset}}.
        \\
        \midrule[1pt]
        \textbf{Propaganda Tactic Detection}&
        Task: Propaganda techniques are methods used in propaganda to convince an audience to believe what the propagandist wants them to believe. Which propaganda techniques does the news contain? Please choose from: {\textit{the candidate label set of a specific dataset}}.
        \\
        \midrule[1pt]
        \textsc{zero-shot}&
        {\textit{Task-related prompt}}\newline
        News: $\boldsymbol{s}$\newline
        Answer:
        \\
        \midrule[1pt]
        \textsc{few-shot}&
        {\textit{Example pairs of news and lable}}\newline
        {\textit{Task-related prompt}}\newline
        News: $\boldsymbol{s}$\newline
        Answer:
        \\
        \midrule[1pt]
        \textsc{retrieval}&
        Knowledge: {\textit{externel knowledge retrieved from Wikipedia}}\newline
        {\textit{Task-related prompt}}\newline
        News: $\boldsymbol{s}$\newline
        Answer:
        \\
        \midrule[1pt]
        \textsc{TAPE}&
        News: $\boldsymbol{s}$\newline
        {\textit{Task-related prompt}}\newline
        Provide your reasoning.\newline
        Answer:
        \\
        \midrule[1pt]
        \textsc{k-hops}&
        News: $\boldsymbol{s}$\newline
        It has the following comments:\newline
        {\textit{the comments related to the news}}\newline
        {\textit{Task-related prompt}}\newline
        Answer:
        \\
        \midrule[1pt]
        \multirow{2}{*}{
        \textsc{k-attention}}&
        News: $\boldsymbol{s}$\newline
        It has the following comments:\newline
        {\textit{the comments related to the news}}\newline
        Task: Please return the index of the comment that best assists in understanding this news.\newline
        Answer:
        \\\cmidrule(r){2-2}
        &
        News: $\boldsymbol{s}$\newline
        It has the following comments:\newline
        {\textit{the selected comments}}\newline
        {\textit{Task-related prompt}}\newline
        Answer:\\
        \bottomrule[1.5pt]
    \end{tabularx}
    \caption{Prompt templates of LLM-based baselines.}
    \label{tab: LLM_baseline_prompts}
\end{table*}

\begin{table*}
    \centering
    \begin{tabularx}{\linewidth}{m{0.3\linewidth}|m{0.65\linewidth}}
        \toprule[1.5pt]
        \multirow{3}{*}{\textsc{F3} \textit{Z-CoT}}&You are an AI assistant trained to detect fake news. Deeply Analyze the given text, think step-by-step, and determine if it is real or fake news. \{News content\}\\\cmidrule(r){2-2}
        &You are an AI assistant trained to detect framing. Deeply Analyze the given text, think step-by-step, which framings does the following news contain? Please choose from: \{label set\} \{News content\}\\\cmidrule(r){2-2}
        &You are an AI assistant trained to detect propaganda tactic. Deeply Analyze the given text, think step-by-step, which propaganda tactics does the following news contain? Please choose from: \{label set\} \{News content\}\\
        \midrule[1pt]
        \multirow{3}{*}{\textsc{F3} \textit{DeF-Gen}}&You are an AI assistant trained to detect fake news. Determine if the text contains factual information supported by evidence (real) or ' 'misleading/inaccurate information (fake). Respond with ’True’ or ’False’. \{News content\}\\\cmidrule(r){2-2}
        &You are an AI assistant trained to detect framing. Framing is a strategic device and a central concept in political communication for representing different salient aspects and perspectives to convey the latent meaning of an issue. Which framings does the following news contain? Please choose from:\{label set\} \{News content\}\\\cmidrule(r){2-2}
        &You are an AI assistant trained to detect propaganda tactic. Propaganda tactics are methods used in propaganda to convince an audience to believe what the propagandist wants them to believe. Which propaganda tactics does the following news contain? Please choose from: \{label set\} \{News content\}\\
        
        \bottomrule[1.5pt]
    \end{tabularx}
    \caption{Prompt templates of LLM-based baselines. (cont.)}
    \label{tab: LLM_baseline_prompts_1}
\end{table*}

\begin{table*}
    \centering
    \begin{tabularx}{\linewidth}{m{0.3\linewidth}|m{0.65\linewidth}}
        \toprule[1.5pt]
        \multicolumn{2}{l}{\parbox{0.98\linewidth}{News:\newline Given the recent wave of allegations pertaining to sexual misconduct on the part of congressmen in Washington, it is not inconceivable that similar accusations could be directed at Senator Chuck Schumer. While I acknowledge that some of my previous posts may have been hasty and not grounded in factual evidence, I must clarify that I do not make such remarks maliciously. I hope this insight assuages any potential doubts about my intentions.}}\\
        \midrule[1pt]
        \multicolumn{2}{l}{\parbox{0.98\linewidth}{User profile:\newline You are a social media user. You are female. You are 18 to 29 years old. Racially, you are White. Financially, your annual family income is less than 30,000. Educationally, you haven't graduated from college. Politically, you are a Democrat. Meanwhile, you are not registered to vote. }}\\
        \midrule[1pt]
        \multicolumn{2}{l}{\parbox{0.98\linewidth}{Comment:\newline As a fellow Democrat, I urge Senator Schumer to address these allegations swiftly and transparently. It's important that we hold all elected officials accountable for their actions.}}\\
        \midrule[1pt]
        \midrule[1pt]
        \textbf{Sentiment Analysis}&
        Based on the content of the news, the three most likely emotions are anger, disgust, and fear. Anger is likely to be felt by those who believe that Senator Schumer has engaged in improper behavior and that he should be held accountable. Disgust is also likely to be felt by those who are repulsed by the idea of sexual misconduct on the part of any politician, regardless of their political affiliation. Fear is likely to be felt by those who are concerned about the potential repercussions of these allegations for the broader political landscape, as well as for the safety and security of the senator himself.
        \\\midrule[1pt]
        \textbf{Framing Detection}&
        The news contains the following five likely framings:\newline 1. Morality: The allegations of sexual misconduct against congressmen in Washington are often framed as a moral issue, with some commentators arguing that such behavior is unethical and unacceptable.\newline 2. Fairness and equality: The issue of sexual misconduct is also framed as a matter of fairness and equality, with some arguing that women should be treated with respect and dignity, regardless of their position or status.\newline 3. Legality, constitutionality and jurisprudence: The legal implications of sexual misconduct are also a key concern, with some commentators arguing that such behavior is a violation of federal law and the Constitution.\newline 4. Policy prescription and evaluation: The issue of sexual misconduct is also framed as a matter of policy prescription and evaluation, with some commentators arguing that Congress should take action to address the problem and prevent it from happening in the future.\newline 5. External regulation and reputation: The issue of sexual misconduct is also framed as a matter of external regulation and reputation, with some commentators arguing that Congress should be held accountable for its actions and that the public should demand better behavior from its elected representatives.
        \\
        \bottomrule[1.5pt]
    \end{tabularx}
    \caption{Examples of the explanations of proxy tasks generated by \ourmethod{}.}
    \label{tab: more_case_1}
\end{table*}

\begin{table*}
    \centering
    \begin{tabularx}{\linewidth}{m{0.3\linewidth}|m{0.65\linewidth}}
        \toprule[1.5pt]
        \textbf{Propaganda Tactics Detection}&
        The news contains the following five propaganda techniques:\newline 1. Appeal to Authority: The news uses the authority of the senator to make the audience believe that he is trustworthy and reliable.\newline 2. Appeal to Popularity: The news uses the popularity of the senator to make the audience believe that he is a good person and deserves their support.\newline 3. Appeal to Fear-Prejudice: The news uses the fear of being accused of sexual misconduct to make the audience believe that the senator is innocent and deserves their support.\newline 4. False Dilemma-No Choice: The news presents the audience with a false dilemma by suggesting that the only choice is to support the senator or to believe the allegations against him.\newline 5. Exaggeration-Minimisation: The news exaggerates the importance of the allegations against the senator and minimizes the evidence supporting them.
        \\\midrule[1pt]
        \textbf{Stance Detection}&
        The stance of sentence 2 on sentence 1 is supportive. The speaker of sentence 2, who is a fellow Democrat, acknowledges the potential allegations against Senator Schumer and urges him to address them swiftly and transparently. This shows that the speaker supports the idea of holding elected officials accountable for their actions, which is a sentiment that is expressed in sentence 1. Additionally, the speaker's use of the word \"fellow\" suggests that they share a common identity with Senator Schumer, further indicating support.
        \\\midrule[1pt]
        \textbf{Response Characterization}&
        Yes, sentence 2 replies to sentence 1. Sentence 1 expresses the possibility that Senator Chuck Schumer could be accused of sexual misconduct, while sentence 2 urges him to address the allegations in a transparent manner. Sentence 1 and Sentence 2 are related in that they both discuss Senator Schumer and the allegations against him.
        \\
        \bottomrule[1.5pt]
    \end{tabularx}
    \caption{Examples of the explanations of proxy tasks generated by \ourmethod{}. (cont.)}
    \label{tab: more_case_2}
\end{table*}

\begin{table*}
    \centering
    \begin{tabularx}{\linewidth}{m{0.3\linewidth}|m{0.65\linewidth}}
        \toprule[1.5pt]
        \textbf{Knowledge Retrieval}&
        Given the recent wave of allegations pertaining to sexual misconduct (Sexual misconduct is misconduct of a sexual nature which exists on a spectrum that may include a broad range of sexual behaviors considered unwelcome. This includes conduct considered inappropriate on an individual or societal basis of morality, sexual harassment and/or criminal sexual assault. However generally, from a purely legal standpoint, sexual misconduct is a \"lay term\" which represents a boundary that has been broken, dictated by a moral set of conduct, particularly where the situation is normally non-sexual and therefore unusual for sexual behavior, or where there is some aspect of personal power or authority that makes sexual behavior inappropriate.) on the part of congressmen (A member of congress (MOC) is a person who has been appointed or elected and inducted into an official body called a congress, typically to represent a particular constituency in a legislature. The term member of parliament (MP) is an equivalent term within a parliamentary system of government.) in Washington (George Washington (February 22, 1732 \u2013 December 14, 1799) was an American Founding Father, military officer, and statesman who served as the first president of the United States from 1789 to 1797. Appointed by the Second Continental Congress as commander of the Continental Army in June 1775, Washington led Patriot forces to victory in the American Revolutionary War and then served as president of the Constitutional Convention in 1787, which drafted and ratified the Constitution of the United States and established the American federal government. Washington has thus been called the \"Father of his Country\".), it is not inconceivable that similar accusations could be directed at Senator (A senate is a deliberative assembly, often the upper house or chamber of a bicameral legislature. The name comes from the ancient Roman Senate (Latin: Senatus), so-called as an assembly of the senior (Latin: senex meaning \"the elder\" or \"old man\") and therefore considered wiser and more experienced members of the society or ruling class. However the Roman Senate was not the ancestor or predecessor of modern parliamentarism in any sense, because the Roman senate was not a de jure legislative body.Many countries have an assembly named a senate, composed of senators who may be elected, appointed, have inherited the title, or gained membership by other methods, depending on the country.) Chuck Schumer (Charles Ellis Schumer ( SHOO-m\u0259r; born November 23, 1950) is an American politician serving as Senate Majority Leader since 2021 and the senior United States senator from New York since 1999. A member of the Democratic Party, he has led the Senate Democratic Caucus since 2017 and was Senate Minority Leader from 2017 to 2021. Schumer is in his fifth Senate term, making him the longest-serving US senator from New York, having surpassed Daniel Patrick Moynihan and Jacob K. Javits in 2023.). While I acknowledge that some of my previous posts may have been hasty and not grounded in factual evidence, I must clarify that I do not make such remarks maliciously. I hope this insight assuages any potential doubts about my intentions.\\
        \bottomrule[1.5pt]
    \end{tabularx}
    \caption{Examples of the explanations of proxy tasks generated by \ourmethod{}. (cont.)}
    \label{tab: more_case_3}
\end{table*}

\end{document}